\documentclass[10pt,journal,compsoc]{IEEEtran}
\ifCLASSOPTIONcompsoc
  \usepackage[nocompress]{cite}
\else
  \usepackage{cite}
\fi
\ifCLASSINFOpdf
  \usepackage[pdftex]{graphicx}
\else
  \usepackage[dvips]{graphicx}
\fi
\usepackage{amsmath}
\usepackage{amssymb}
\usepackage{bbding}

\usepackage{color}
\newcommand{\blue}[1]{\textcolor{black}{#1}}

\definecolor{skyblue}{RGB}{166, 206, 227}
\definecolor{rosy}{RGB}{251, 154, 153}
\definecolor{redline}{RGB}{227, 26, 28}

\def\eg{\emph{e.g.}}
\def\ie{\emph{i.e.}}

\begin{document}
%
\title{Human-Centric Transformer for \\
Domain Adaptive Action Recognition}
%
%
%
%

\author{Kun-Yu Lin, Jiaming Zhou, Wei-Shi Zheng
\IEEEcompsocitemizethanks{\IEEEcompsocthanksitem Kun-Yu Lin and Jiaming Zhou are with the School of Computer Science and Engineering, Sun Yat-sen University, Guangzhou 510275, China,
the Key Laboratory of Machine Intelligence and Advanced Computing, Ministry of Education, Guangzhou 510275, China,
and also with the Guangdong Province Key Laboratory of Information Security Technology, Sun Yat-sen University, Guangzhou 510275, China.
\protect\\
E-mail: \{linky5, zhoujm55\}@mail2.sysu.edu.cn.
\IEEEcompsocthanksitem Wei-Shi Zheng is with the School of Computer Science and Engineering, Sun Yat-sen University, Guangzhou 510275, China,
and also with the Peng Cheng Laboratory, Shenzhen 518005, China.\protect\\
E-mail: wszheng@ieee.org.
\IEEEcompsocthanksitem Corresponding author: Wei-Shi Zheng.
}
}

%
%

\markboth{SUBMISSION TO IEEE TRANSACTIONS ON PATTERN ANALYSIS AND MACHINE INTELLIGENCE}%
{Shell \MakeLowercase{\textit{et al.}}: Bare Demo of IEEEtran.cls for Computer Society Journals}
%



\IEEEtitleabstractindextext{%
\begin{abstract}
We study the domain adaptation task for action recognition, namely domain adaptive action recognition, which aims to effectively transfer action recognition power from a label-sufficient source domain to a label-free target domain.
Since actions are performed by humans, it is crucial to exploit human cues in videos when recognizing actions across domains.
However, existing methods are prone to losing human cues but prefer to exploit the correlation between non-human contexts and associated actions for recognition, and the contexts of interest agnostic to actions would reduce recognition performance in the target domain.
To overcome this problem, we focus on uncovering human-centric action cues for domain adaptive action recognition, and our conception is to investigate two aspects of human-centric action cues, namely human cues and human-context interaction cues.
Accordingly, our proposed \textit{Human-Centric Transformer (HCTransformer)} develops a decoupled human-centric learning paradigm to explicitly concentrate on human-centric action cues in domain-variant video feature learning.
Our HCTransformer first conducts human-aware temporal modeling by a human encoder, aiming to avoid a loss of human cues during domain-invariant video feature learning.
Then, by a Transformer-like architecture, HCTransformer exploits domain-invariant and action-correlated contexts by a context encoder, and further models domain-invariant interaction between humans and action-correlated contexts.
We conduct extensive experiments on three benchmarks, namely UCF-HMDB, Kinetics-NecDrone and EPIC-Kitchens-UDA, and the state-of-the-art performance demonstrates the effectiveness of our proposed HCTransformer.
\end{abstract}

\begin{IEEEkeywords}
Action recognition, domain adaptive action recognition.
\end{IEEEkeywords}}

\maketitle

\IEEEdisplaynontitleabstractindextext

%
\IEEEpeerreviewmaketitle

\begin{figure}[t]
\vskip -0.2in
\begin{center}
    \centering
    \includegraphics[width=0.92\linewidth]{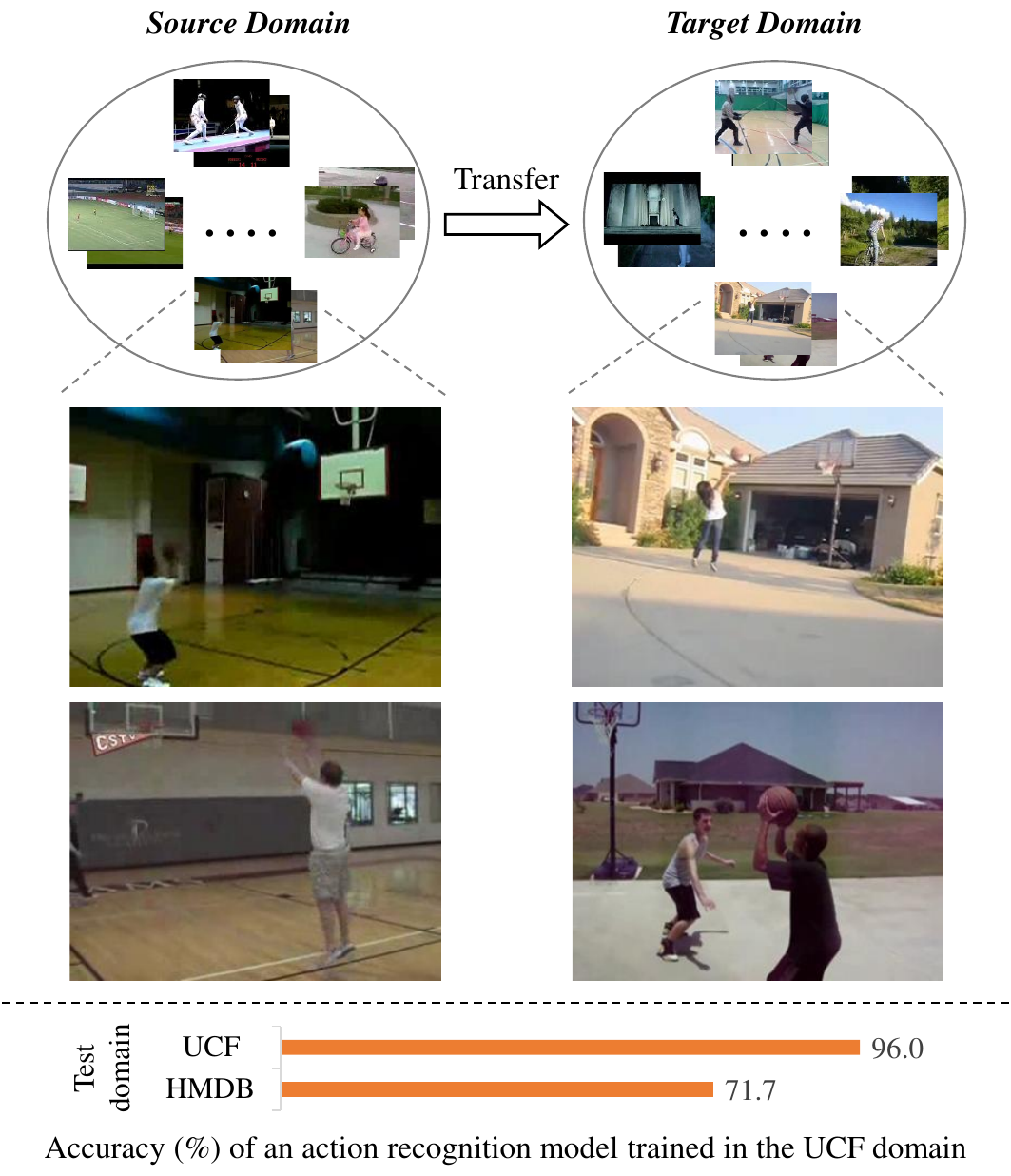}
\end{center}
\vskip -0.15in
\caption{
Top: An illustration of domain adaptive action recognition.
We zoom in example videos of the action ``shoot ball'' for clarity.
As shown in the figure, videos of the same action look very different in the source and target domains, due to environment change, viewpoint diversity, illumination shift, etc.
Bottom: The performance of an action recognition model trained in the UCF domain drops significantly (24.3\% in terms of accuracy) when testing in the HMDB domain.
Best viewed in color.}
\vskip -0.1in
\label{fig:setting}
\end{figure}

\IEEEraisesectionheading{\section{Introduction}\label{sec:introduction}}
\IEEEPARstart{A}{ction}
recognition aims to recognize what actions humans are performing in videos, which has wide applications in sport analysis, health monitoring, surveillance systems, robotics and autonomous driving, etc~\cite{DBLP:journals/ijcv/KongF22,9795869,DBLP:conf/iccv/DingLH0TB23,DBLP:journals/corr/abs-2406-08877}.
Recently, many advanced network architectures have been proposed for action recognition~\cite{DBLP:conf/iccv/TranBFTP15,DBLP:conf/cvpr/CarreiraZ17,DBLP:conf/cvpr/TranWTRLP18,DBLP:conf/eccv/WangXW0LTG16,DBLP:conf/eccv/ZhouAOT18,DBLP:conf/iccv/LinGH19,DBLP:conf/cvpr/HusseinGS19,DBLP:conf/cvpr/ZhouLLZ21}.
These action recognition models usually assume similar test scenarios to the training ones, \ie, it implicitly assumes that the training and test videos follow the same distribution~\cite{DBLP:conf/cvpr/SultaniS14,DBLP:journals/ivc/XuZWF16,DBLP:conf/bmvc/JamalNDV18,DBLP:conf/iccv/ChenKAYCZ19,DBLP:conf/cvpr/KarpathyTSLSF14,DBLP:conf/cvpr/CarreiraZ17,DBLP:journals/corr/Abu-El-HaijaKLN16}. Thus, these models can leverage abundant cues in videos (\eg, human bodies, action-correlated contexts) for recognizing actions.
However, this assumption is too rigorous in practice, since machine intelligence systems often face unfamiliar scenarios when deployed in real-world applications.
For example, a domestic robot will work in a new house, and a surveillance system will encounter illumination shift caused by camera viewpoint, light color, or weather change \cite{Volpi_2022_CVPR,DBLP:conf/iccv/LengyelGMG21,xu2020arid}.
As a result, when deployed in unfamiliar scenarios, action recognition models would suffer from a significant performance drop, as shown in the bottom of Figure~\ref{fig:setting}.

\begin{figure*}[t]
\begin{center}
    \centering
    \includegraphics[width=0.95\linewidth]{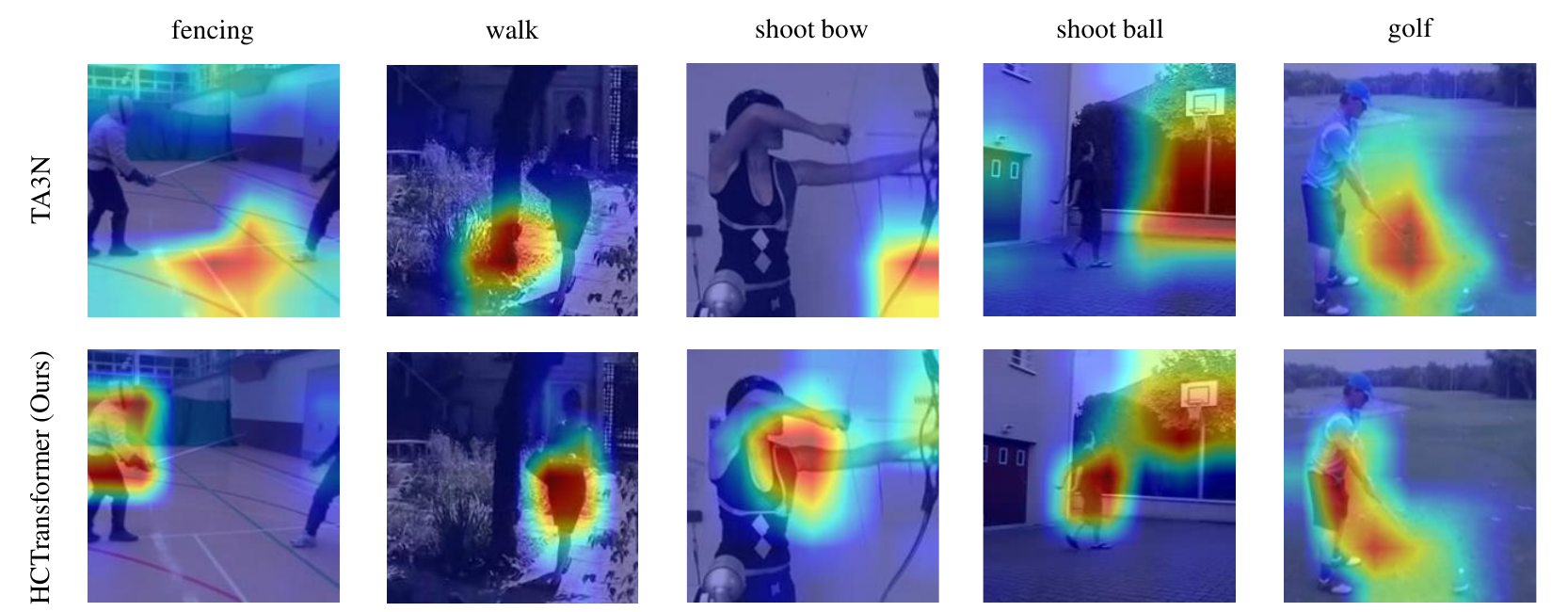}
\end{center}
\vskip -0.1in
\caption{Existing methods based on exploring domain invariance (\eg, TA3N \cite{DBLP:conf/iccv/ChenKAYCZ19}) are prone to lose their focus on human cues in videos. We demonstrate Grad-CAM~\cite{DBLP:journals/ijcv/SelvarajuCDVPB20} examples about the problem.
As shown by the heatmaps, TA3N focuses on non-human context cues in videos. In some cases, contexts of interest are agnostic to the performing actions, \eg, floor of the court in the ``fencing'' video, which results in recognition errors. In contrast to TA3N, our proposed HCTransformer focuses on human-centric action cues closely related to the performing actions, \eg, the fencer's body parts in the ``fencing'' video.
Videos in the figure are from the target domain of UCF-HMDB. Best viewed in color.
Please refer to Figure~\ref{fig:more_baselines} for the Grad-CAM visualization of more existing domain adaptive action recognition methods.
}
\label{fig:intro}
\end{figure*}

Domain adaptive action recognition can alleviate the above problem~\cite{DBLP:conf/cvpr/SultaniS14,DBLP:journals/ivc/XuZWF16,DBLP:conf/bmvc/JamalNDV18,DBLP:conf/iccv/ChenKAYCZ19}.
This task aims to effectively transfer action recognition power from a label-sufficient source domain to a label-free target domain, where videos from the source and target domains follow different distributions but share an identical label space.
In this task, videos of the same action are visually different across domains,
\eg, humans in the source domain perform basketball shooting in indoor basketball courts while humans in the target domain perform outdoors
as shown in the top of Figure~\ref{fig:setting}.
In contrast to the traditional image-based domain adaptation task \cite{DBLP:conf/icml/LongC0J15,DBLP:journals/corr/TzengHZSD14,DBLP:journals/jmlr/GaninUAGLLML16}, domain adaptive action recognition aims at recognizing dynamic human actions performed in videos rather than static objects or scenes in images, and therefore complex and diverse motion cues should be taken into account beyond appearance cues.

In the absence of video labels in the target domain, a typical way to tackle domain adaptive action recognition is exploring the invariance (\ie, shared information) between videos from the source and target domains.
Accordingly, existing domain adaptive action recognition methods mainly consist of two components, namely temporal modeling modules for video feature extraction and domain alignment losses for domain-invariant feature learning~\cite{DBLP:conf/iccv/ChenKAYCZ19,DBLP:conf/aaai/PanCAN20,DBLP:conf/mm/LuoHW0B20,xu2022aligning}.
For example, TA3N~\cite{DBLP:conf/iccv/ChenKAYCZ19} explores domain-invariant features encoding temporal dynamics based on domain discrepancy; TCoN~\cite{DBLP:conf/aaai/PanCAN20} selects critical domain-invariant clips in videos by using cross-domain co-attention modules;
ABG~\cite{DBLP:conf/mm/LuoHW0B20} proposes to model cross-domain information interaction for domain-invariant feature learning, which constructs a bipartite graph composed of heterogeneous vertexes for graphical feature learning;
ACAN~\cite{xu2022aligning} proposes to learn domain-invariant dependencies among pixels across spatiotemporal dimensions.

Prevailing methods based on domain invariance are prone to losing human cues and concentrate on non-human context cues when recognizing actions across domains, which attributes to the strong correlation between the contexts and associated actions within a single domain.
As shown in Figure~\ref{fig:intro}, the prevailing TA3N \cite{DBLP:conf/iccv/ChenKAYCZ19} prefers to concentrate on non-human objects or scenes, with little attention on humans.
Unfortunately, the contexts of interest agnostic to the performing actions would cause recognition errors.
It is because contexts are vague for action recognition (also known as the spurious correlation~\cite{DBLP:conf/eccv/LiLV18,DBLP:conf/nips/ChoiGMH19,DBLP:conf/cvpr/WangGLLM0PHJS21,DBLP:journals/ijcv/WeinzaepfelR21,DBLP:conf/iccv/LiLZL23}), that is an action could correlate with different types of contexts in different domains.
Therefore, action-agnostic contexts in the target domain would largely reduce the action recognition performance, although these contexts strongly correlate with actions in the source domain.
For example, as shown in the top of Figure~\ref{fig:setting}, if humans usually play ball shooting in indoor basketball courts in the source domain, previous models would tend to recognize the action ``shoot ball'' by recognizing the courts.
However, in the target domain, humans shoot ball outdoors and fence in indoor courts at the same time, thus recognizing only the courts would lead to wrong predictions in this case (see the first column of Figure~\ref{fig:intro}).

To address the above problem, it is crucial to concentrate on human-centric action cues when recognizing actions across domains, since actions are performed by humans.
However, due to the abundant contents in videos and a limited amount of labeled training data, it is challenging to accomplish this idea with only domain alignment loss guidance on top of video feature extraction modules as previous works (\eg, TA3N~\cite{DBLP:conf/iccv/ChenKAYCZ19} applies domain alignment losses at several feature levels).
To solve this challenge, we consider that the human-centric action cues in videos should consist of two aspects, namely human cues and human-context interaction cues.
The human cues refer to appearance or motion cues of action performers, and the human-context interaction cues refer to the interaction between action performers and action-correlated contexts (\eg, objects, scenes).
Accordingly, we propose to explicitly decouple the video feature extraction process into two parts, namely human-aware modeling and human-context interaction modeling.
By such modeling, we are able to explicitly constrain the model to concentrate on the human-centric action cues in domain-invariant video feature learning.
For example, as shown in Figure~\ref{fig:intro}, our proposed model concentrates on human-centric cues when recognizing the action ``shoot bow'', \ie, we focus on not only the body parts of the archer but also the interaction between the archer and bow, which is robust to domain shift.

To \textit{explicitly} exploit domain-invariant human-centric action cues, we propose \textit{Human-Centric Transformer (HCTransformer)}, which follows a novel decoupled human-centric learning paradigm.
Specifically, our proposed HCTransformer simultaneously performs temporal modeling and cross-domain feature alignment at three semantic levels, namely human, context and human-context interaction.
Firstly, HCTransformer conducts human-aware temporal modeling by a human encoder, which constrains the model to concentrate on human cues in domain-invariant video feature learning.
Secondly, we introduce a context encoder to exploit domain-invariant and action-correlated contexts in non-human parts of videos, which leverages layer-by-layer attention mechanisms with context prototypes as extra tokens.
Furthermore, we introduce a human-context decoder to model domain-invariant interaction between humans and contexts, leveraging a query-based architecture resembling Transformers.

Our proposed HCTransformer achieves state-of-the-art performance on three benchmarks, namely UCF-HMDB, Kinetics-NecDrone and EPIC-Kitchens-UDA.
The superior performance on both human-centric and hand-centric benchmarks demonstrate the versatility of HCTransformer.
Also, our qualitative and quantitative
analysis shows that HCTransformer prefers to focus on human-centric action cues compared with existing methods (\eg, Figure~\ref{fig:more_baselines}).
Overall, the extensive quantitative and qualitative experimental results demonstrate the effectiveness of our proposed HCTransformer.

\section{Related Works}
\label{sec:related}

\subsection{Human Activity Understanding}
Action recognition aims to recognize human actions in videos. Recently, with the success of deep learning~\cite{DBLP:conf/nips/KrizhevskySH12,DBLP:journals/corr/SimonyanZ14a,DBLP:conf/cvpr/SzegedyLJSRAEVR15,DBLP:conf/cvpr/HeZRS16,DBLP:conf/nips/SutskeverVL14,DBLP:conf/emnlp/ChoMGBBSB14,DBLP:conf/nips/VaswaniSPUJGKP17,DBLP:conf/iccv/TangP023,DBLP:journals/tmm/JiaoTLGMWZ23,DBLP:journals/pami/PengLZ24}, many advanced video classification architectures have been proposed.
Considering the temporal dimension, one class of representative works extends 2D convolution for image processing to 3D convolution for video processing \cite{DBLP:conf/iccv/TranBFTP15,DBLP:conf/cvpr/CarreiraZ17,DBLP:conf/cvpr/TranWTRLP18,DBLP:conf/iccv/TranWFT19,DBLP:conf/cvpr/Feichtenhofer20,DBLP:conf/iclr/ZhangGHS020,DBLP:conf/iclr/LiLWWQ21}. Another class of works applies 2D convolution for frame-level spatial modeling and conducts spatial-temporal modeling after frame-level feature extraction~\cite{DBLP:conf/eccv/WangXW0LTG16,DBLP:conf/eccv/ZhouAOT18}.
Based on 2D convolution, some works propose to couple explicit shifts along the time dimension for efficient temporal modeling \cite{DBLP:conf/iccv/LinGH19,DBLP:conf/aaai/ShaoQL20,DBLP:conf/cvpr/SudhakaranEL20}. Recently, motivated by the success of Vision Transformer \cite{DBLP:conf/iclr/DosovitskiyB0WZ21,DBLP:conf/iccv/LiuL00W0LG21}, pioneer works have made some efforts in adapting Transformer~\cite{DBLP:conf/nips/VaswaniSPUJGKP17} for action recognition~\cite{DBLP:conf/cvpr/GirdharCDZ19,DBLP:journals/corr/abs-2102-00719,DBLP:conf/mm/ZhangHN21,DBLP:conf/iccv/ZhangLLSZBCMT21,DBLP:conf/icml/BertasiusWT21,DBLP:conf/iccv/Arnab0H0LS21}.

Although these advanced video classification architectures achieve appealing performance, they usually require a large number of labeled action videos for training, which costs expensive labors for annotations~\cite{DBLP:conf/cvpr/KarpathyTSLSF14,DBLP:conf/cvpr/CarreiraZ17,DBLP:journals/corr/Abu-El-HaijaKLN16}. What is more, the above models usually assume an identical test distribution to the training one, which is not practical in real-world applications.
In comparison, our work focuses on a domain adaptation setting for action recognition, where the distribution shift between the training and test phases is taken into account.
To tackle domain adaptive action recognition, our work focuses on a specific problem of this task, \ie, previous methods based on domain invariance are prone to losing human cues and concentrating on non-human contexts.
And accordingly, we propose a decoupled human-centric learning paradigm to specifically address this problem.
Our proposed HCTransformer first constrains the human encoder to concentrate on human cues and then uses the human-context decoder to model domain-invariant human-context interaction cues beyond human cues, which significantly alleviates the misguidance of non-human contexts in domain adaptive action recognition.

Human-object interaction detection is another fundamental task for human activity understanding, which aims to localize pairs of human and object instances and recognize the interactions between them~\cite{DBLP:journals/corr/GuptaM15,DBLP:conf/cvpr/GkioxariGDH18,DBLP:conf/cvpr/MoraisLV021,DBLP:conf/iccv/JiDN21}. Different from action recognition, human-object interaction detection is technically based on detection rather than classification, where human and object bounding boxes should be involved as annotations in training. Existing works perform human-object interaction detection mainly in still images~\cite{DBLP:journals/corr/GuptaM15,DBLP:conf/cvpr/GkioxariGDH18}.
Usually, two-stage methods follow a paradigm that first produces human and object proposals by a pre-trained object detector and then predicts the interactions for each human-object pair \cite{DBLP:conf/cvpr/GkioxariGDH18,DBLP:conf/bmvc/GaoZH18,DBLP:conf/eccv/QiWJSZ18,DBLP:conf/eccv/GaoXZH20}.
In addition, one-stage methods are proposed to address the feature mismatch problem in two-stage methods, which also leads to better efficiency \cite{DBLP:conf/cvpr/LiaoLWCQF20,DBLP:conf/cvpr/WangYDK0S20}. Recently, some pioneer works have explored advanced Transformer-based architectures beyond one-stage methods and achieved significant performance improvement \cite{DBLP:conf/cvpr/KimLKKK21,DBLP:conf/cvpr/ZouWHLWZLZZW021,DBLP:conf/cvpr/TamuraOY21,DBLP:conf/cvpr/DongLXZY0Z22}.
In the next subsection, we will review the domain adaptive action recognition task, a cross-domain action classification task where bounding box annotations are inaccessible for training.

\subsection{Domain Adaptive Action Recognition}
Domain adaptive action recognition is a domain adaptation task in the action recognition field, which aims to transfer action recognition power from a label-sufficient source domain to a label-free target domain.
Existing works mainly focus on exploring domain invariance.
TA3N~\cite{DBLP:conf/iccv/ChenKAYCZ19} and TCoN~\cite{DBLP:conf/aaai/PanCAN20} focus on critical domain-invariant clips in a video by using different attention mechanisms. Specifically, TA3N \cite{DBLP:conf/iccv/ChenKAYCZ19} leverages the entropy of domain discriminators, and TCoN \cite{DBLP:conf/aaai/PanCAN20} leverages the cross-domain feature similarity.
ACAN~\cite{xu2022aligning} models the dependencies among pixels across spatiotemporal dimensions when exploring domain invariance, and ABG~\cite{DBLP:conf/mm/LuoHW0B20} constructs heterogeneous graphs and models cross-domain information interaction for domain-invariant feature learning.
Besides, SAVA \cite{DBLP:conf/eccv/ChoiSSH20} and CoMix \cite{DBLP:conf/nips/SahooSPSD21} borrow ideas from the self-supervised literature for filtering out domain-specific information. SAVA \cite{DBLP:conf/eccv/ChoiSSH20} leverages the clip order prediction task, while CoMix \cite{DBLP:conf/nips/SahooSPSD21} leverages temporal contrastive losses with background mixing.

Beyond a single RGB modality, some works \cite{DBLP:conf/cvpr/MunroD20,DBLP:conf/cvpr/SongZ0Y0HC21,DBLP:conf/iccv/KimTZ0SSC21,DBLP:conf/cvpr/YangHSS22,Zhang_2022_CVPR} explore the multi-modality nature of videos.
MM-SADA \cite{DBLP:conf/cvpr/MunroD20} introduces a cross-modality consistency constraint for cross-modal alignment. Song et al. \cite{DBLP:conf/cvpr/SongZ0Y0HC21} and Kim et al. \cite{DBLP:conf/iccv/KimTZ0SSC21} exploit the contrastive loss in multi-modal scenarios for domain alignment.
Yang et al. \cite{DBLP:conf/cvpr/YangHSS22} consider cross-modal complementarity and highlight the most transferrable aspects by cross-modal consensus. Considering further the audio modality, Zhang et al. \cite{Zhang_2022_CVPR} propose to guide visual feature learning by excluding actions with distinctive sounds.

Existing domain adaptive action recognition methods usually explore domain invariance according to the guidance of domain alignment losses or contrastive losses, and they do not explicitly concentrate on human cues in video feature learning.
As a result, these methods are prone to losing human cues and prefer to recognize actions by the correlation between contexts and actions, thus resulting in recognition errors since actions correlate with contexts differently across domains.
Our work forces the model to concentrate on human-centric action cues by a novel decoupled human-centric learning paradigm.

Some pioneer works~\cite{DBLP:journals/pami/YaoWWYL22,DBLP:conf/wacv/PlanamentePAC22,DBLP:journals/corr/abs-2310-17942,DBLP:conf/iccv/PlizzariPCD23,DBLP:journals/corr/abs-2403-01560} explore a domain generalization setting for action recognition, which aims at learning generalizable representations without access to the target domain.
And, some other works focus on variant domain adaptation settings of action recognition, such as partial~\cite{DBLP:conf/iccv/XuYCCLM21}, open-set~\cite{DBLP:journals/pami/BustoIG20} and source-free~\cite{DBLP:conf/eccv/XuYCWWC22} settings.

\begin{figure*}[t]
\begin{center}
    \includegraphics[width=0.95\linewidth]{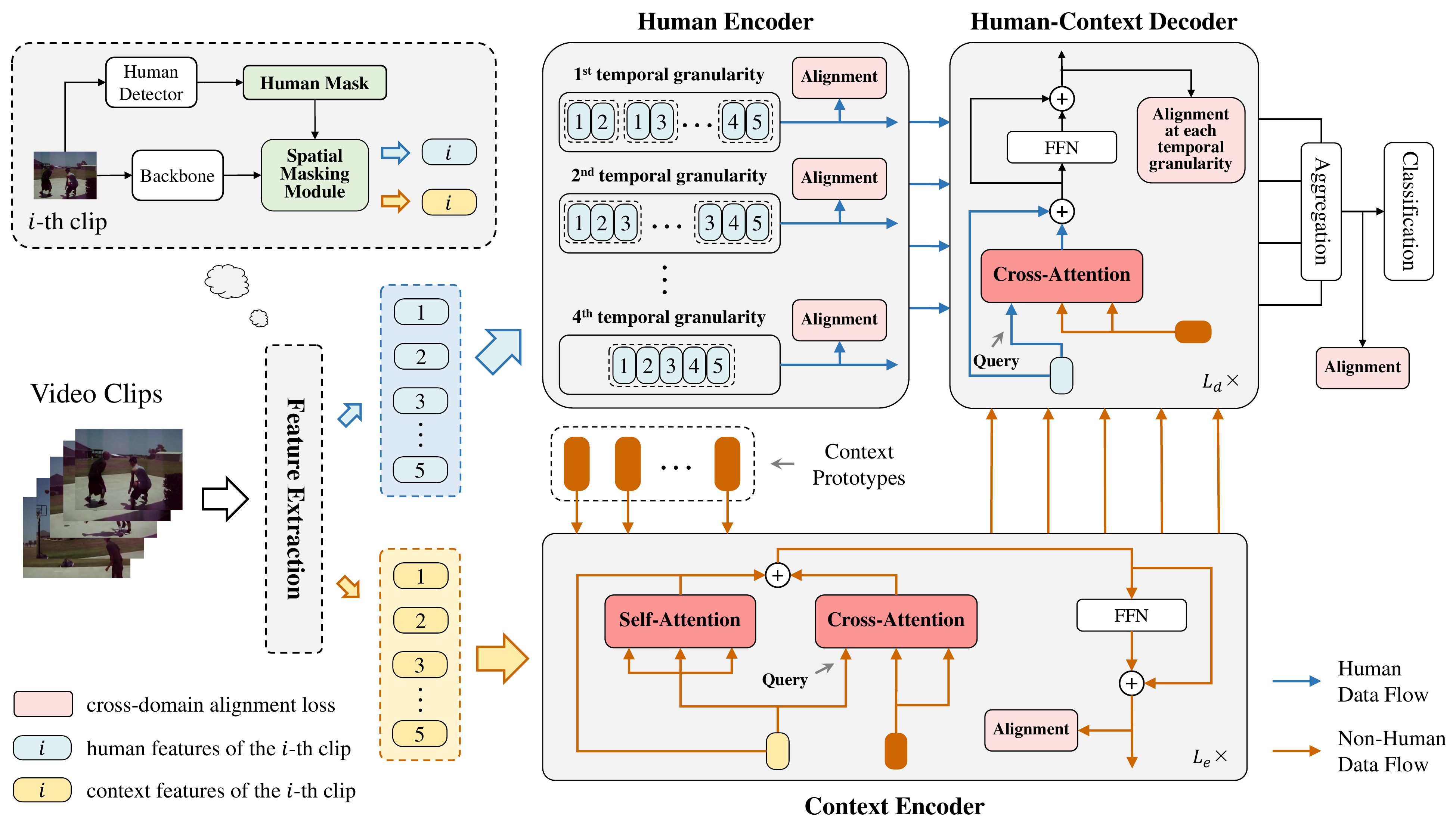}
\end{center}
\caption{An overview of the proposed Human-Centric Transformer (HCTransformer). We use a five-clip video for demonstration (\ie, $M=5$). HCTransformer mainly consists of three components, namely human encoder, context encoder and human-context decoder, aiming at learning human-centric action cues and aligning feature distributions at different levels. The human encoder focuses on temporal modeling for human cues, where feature alignment is conducted at each temporal granularity respectively. By introducing context prototypes as extra tokens, the context encoder exploits domain-invariant and action-correlated contexts in non-human parts of videos using self-attention and cross-attention modules. By taking the outputs of the two encoders as inputs, the human-context decoder further models the interaction between humans and action-correlated contexts using cross-attention modules. Best viewed in color.}
\label{fig:overview}
\end{figure*}

\subsection{General Domain Adaptation}
Domain adaptation is one of the most classical transfer learning tasks, where the source and target domains follow different but related distributions with identical label space \cite{DBLP:journals/tkde/PanY10}.
This task aims at the generalization in a label-free target domain, by means of transferring knowledge from a label-sufficient source domain to this target domain.
Given labeled samples in the source domain and unlabeled samples in the target domain for training, domain adaptation attempts to learn transferable models by mitigating the distribution shift.

Typically, existing methods focus on exploring domain invariance.
According to the theoretical upper bound \cite{DBLP:journals/ml/Ben-DavidBCKPV10}, some works propose to minimize specific statistical distances (\eg,  maximum mean discrepancy) between the source and target domains \cite{DBLP:conf/icml/LongC0J15,DBLP:conf/icml/LongZ0J17,DBLP:journals/corr/TzengHZSD14,DBLP:conf/eccv/DamodaranKFTC18,DBLP:conf/eccv/SunS16}. Inspired by generative adversarial networks \cite{DBLP:conf/nips/GoodfellowPMXWOCB14}, adversarial-learning-based methods aim to extract domain-invariant features such that auxiliary domain discriminators cannot distinguish samples from different domains \cite{DBLP:journals/jmlr/GaninUAGLLML16,DBLP:conf/cvpr/SaitoWUH18,DBLP:conf/cvpr/TzengHSD17,DBLP:conf/icml/0002LLJ19}. Besides, some works focus on some specific differences between domains for implicit domain alignment, \eg, the scale of feature norm \cite{DBLP:conf/iccv/XuLYL19}, class confusion \cite{DBLP:conf/eccv/JinWLW20}. What is more, by assigning pseudo labels to unlabeled training samples, pseudo-label-based methods have the potential to exploit domain-specific information beyond domain-invariant information~\cite{DBLP:conf/icml/SaitoUH17,DBLP:conf/cvpr/Tang0J20,DBLP:conf/cvpr/LiangHF21}.
Some other works focus on variant domain adaptation settings, such as partial~\cite{DBLP:conf/cvpr/CaoL0J18,DBLP:conf/eccv/LinZQZ22}, open-set~\cite{DBLP:conf/iccv/BustoG17,DBLP:conf/eccv/SaitoYUH18} and universal~\cite{DBLP:conf/cvpr/YouLCWJ19,DBLP:conf/cvpr/0005KZW021} settings.

Traditional domain adaptation usually focuses on recognizing objects in images, while domain adaptive action recognition focuses on recognizing dynamic human actions in videos. With the extra temporal dimension, videos are more complex than individual images, and thus multifarious temporal modeling modules are usually involved for video feature extraction.
Our work contributes a novel video architecture named HCTransformer for domain adaptive action recognition, which explicitly decouples the video feature extraction process to concentrate on human-centric action cues.

\section{Human-Centric Transformer}
\label{sec:method}
In this section, we elaborate on the proposed Human-Centric Transformer (HCTransformer), which exploits a decoupled human-centric learning paradigm to learn domain-invariant human-centric action cues for domain adaptive action recognition.

\subsection{Problem Formulation}
\label{subsec:formulation}
In domain adaptive action recognition, there are a set of labeled videos $\mathcal{D}_s=\{(F^s, y^s)\}$ from the source domain and a set of unlabeled videos $\mathcal{D}_t=\{F^t\}$ from the target domain, which consist of $N_s$ and $N_t$ videos, respectively.
$F^s$/$F^t$ denotes a video from the source/target domain, and $y^s$ denotes the label of video $F^s$.
The goal of domain adaptive action recognition is learning a transferable model to predict action labels in the label-free target domain. In domain adaptive action recognition, the source video set $\{F^s\}$ and target video set $\{F^t\}$ follow different but related distributions. The source and target domains share the same label space ${\mathcal{Y}\in\mathbb{R}^{N_{\mathrm{cls}}}}$, where $N_{\mathrm{cls}}$ denotes the number of action classes. Following the standard domain adaptive action recognition setting \cite{DBLP:conf/iccv/ChenKAYCZ19}, in our work, each video is evenly divided into $M$ segments, \ie, $F^s=\{f^s_{i}\}^{M}_{i=1}, F^t=\{f^t_{i}\}^{M}_{i=1}$ and $i$ indicates the segment index. A clip of frames is sampled from each segment as model input during training and testing. In the following illustration, we omit the superscript ``s'' or ``t'' to avoid cluttered notations.

\subsection{Decoupled Human-Centric Learning Paradigm}
\label{subsec:overview}
Since actions are performed by humans with certain contexts (\eg, tools), we conceive that human-centric cues for recognizing actions across domains consist of two parts, namely the cues of action performers and the interaction between action performers and action-correlated contexts. Accordingly, in our proposed \textit{Human-Centric Transformer (HCTransformer)}, we develop a decoupled human-centric learning paradigm, which explicitly concentrates on human-centric action cues in domain-invariant video feature learning. Our HCTransformer simultaneously conducts temporal modeling and cross-domain feature alignment at three semantic levels, namely human, context and human-context interaction. An overview of the proposed HCTransformer is shown in Figure~\ref{fig:overview}. In what follows, we illustrate the overall decoupled human-centric learning paradigm of our proposed HCTransformer.

First of all, given a video $F$, HCTransformer takes $M$ video clips as inputs and extracts $M$ clip-level feature maps using the backbone (\eg, ResNet~\cite{DBLP:conf/cvpr/HeZRS16}, I3D \cite{DBLP:conf/cvpr/CarreiraZ17}). For a 3D backbone (\eg, I3D), we reduce the temporal dimension of each feature map to one by average pooling.
The $M$ feature maps are denoted by $X=[x_{1}, x_{2}, ..., x_{M}]^T\in\mathbb{R}^{M\times H\times W\times D}$, where $D$ denotes the feature dimension, $H\times W$ denotes the size of feature maps and $x_i\in\mathbb{R}^{H\times W\times D}$ denotes the feature map for the $i$-th clip.

After clip-level feature extraction, HCTransformer divides the pixels of each clip-level feature map into two sets by a spatial masking module.
Specifically, based on the original frame (\ie, the center frame of the clip) corresponding to the feature map $x_i$, we obtain a human mask by an off-the-self human detector and downsample the mask to the size of feature maps.
The downsampled human mask is denoted by $m_i\in\mathbb{R}^{H\times W}$, where each item of $m_i$ indicates the probability of visible humans in the corresponding position. If the probability is larger than the threshold $\lambda_m$, we regard that the corresponding pixels belong to human parts. In this way, the spatial masking module produces two sets of features for each clip, representing human and non-human context parts, respectively. For the $i$-th clip, the human feature set is denoted by $x_i^{\mathrm{hm}}$ and the non-human context feature set is denoted by $x_i^{\mathrm{ctx}}$.

Based on the clip-level human and non-human context features, HCTransformer conducts temporal modeling simultaneously with cross-domain feature alignment from three perspectives, namely human, context and human-context interaction. Accordingly, HCTransformer consists of three components, namely human encoder, context encoder and human-context decoder.
Specifically, the overall structure of HCTransformer is formulated as follows:
\begin{equation}
\begin{aligned}
    Z^{\mathrm{hm}}   & = \mathrm{HmEnc}(X^{\mathrm{hm}}),     \\
    Z^{\mathrm{ctx}}  & = \mathrm{CtxEnc}(X^{\mathrm{ctx}}, C^{\mathrm{ctx}}),     \\
    Z^{\mathrm{hc}}   & = \mathrm{HCDec}(Z^{\mathrm{hm}}, Z^{\mathrm{ctx}}),
    \label{eq:overall}
\end{aligned}
\end{equation}
where $X^{\mathrm{hm}}=[x_1^{\mathrm{hm}}, x_2^{\mathrm{hm}}, ..., x_M^{\mathrm{hm}}]^T$ denotes clip-level human features and $X^{\mathrm{ctx}}=[x_1^{\mathrm{ctx}}, x_2^{\mathrm{ctx}}, ..., x_M^{\mathrm{ctx}}]^T$ denotes clip-level context features in the video.
The human encoder $\mathrm{HmEnc}(\cdot)$ focuses on temporal modeling for human cues, where feature alignment is conducted at different temporal granularities.
Since videos contain abundant contexts, the context encoder $\mathrm{CtxEnc}(\cdot)$ is proposed to learn domain-invariant and action-correlated contexts in non-human parts of videos by introducing context prototypes $C^{\mathrm{ctx}}$ and attention mechanisms.
Taking the outputs of the two encoders as inputs, the human-context decoder $\mathrm{HCDec}(\cdot)$ further models domain-invariant interaction between humans and contexts for domain adaptive action recognition.
In what follows, we illustrate these three components of our proposed HCTransformer in detail.

\subsection{Human-Aware Temporal Modeling}
\label{subsec:hm-enc}
Our proposed human encoder $\mathrm{HmEnc}(\cdot)$ performs temporal modeling for cues of action performers, namely human-aware temporal modeling.
The goal of our human encoder is to learn domain-invariant video-level human features, which are robust to domain shifts when recognizing actions across domains.

Specifically, for the $i$-th clip, the set of clip-level human features is denoted by $x_i^{\mathrm{hm}} = \{x_{i,j} | m_{i,j}\geq\lambda_m, j=1,2,\dots,HW\}$.
Since there could be other persons except action performers in videos, we introduce a parametric module to aggregate human features $x_i^{\mathrm{hm}}$ as human encoder inputs.
The input of the human encoder is formulated as follows:
\begin{equation}
\begin{aligned}
    \hat{x}_i^{\mathrm{hm}} = \mathrm{AvgPool}\left(\mathrm{SelfAttn}(x_i^{\mathrm{hm}})\right),
    \label{eq:hm-input}
\end{aligned}
\end{equation}
where $\mathrm{AvgPool}(\cdot)$ denotes the average pooling operation and $\mathrm{SelfAttn}(\cdot)$ denotes a self-attention module.
By introducing such a parametric module, our model tends to focus on the action performers instead of all persons (\ie, ignore persons irrelevant to the actions being performed) according to the guidance of the action classification loss. Average pooling is applied to aggregate multiple features into one.

Based on input features $\{\hat{x}_i^{\mathrm{hm}}\}$, we figure out the domain-invariant temporal dynamics for human cues. Since videos contain complex temporal dynamics, we prefer to align feature distributions at each temporal granularity in temporal modeling for better domain alignment, inspired by previous works \cite{DBLP:conf/icml/LongC0J15,DBLP:conf/iccv/ChenKAYCZ19}. To this end, we adopt the Temporal Relation Network (TRN) \cite{DBLP:conf/eccv/ZhouAOT18} for temporal modeling, which extracts $M-1$ orders of temporal dynamics features explicitly corresponding to $M-1$ temporal granularities. Specifically, in our human encoder, the $i$-th TRN module aims at learning features at the $i$-th temporal granularity, which models the temporal information interaction among $i+1$ sampled clips.
We denote the human temporal features by $Z^{\mathrm{hm}} = [z_1^{\mathrm{hm}}, z_2^{\mathrm{hm}}, ..., z_{M-1}^{\mathrm{hm}}]^T\in\mathbb{R}^{(M-1)\times D_v}$, where $D_v$ is the feature dimension. $z_i^{\mathrm{hm}}$ denotes the feature of the $(i+1)$-order temporal dynamics, which is given as follows:
\begin{equation}
\begin{aligned}
    z_i^{\mathrm{hm}} = \mathrm{TRN}_i\left(\mathrm{Proj}(\hat{x}_1^{\mathrm{hm}}), \mathrm{Proj}(\hat{x}_2^{\mathrm{hm}}), ..., \mathrm{Proj}(\hat{x}_M^{\mathrm{hm}})\right),
    \label{eq:hm-trn}
\end{aligned}
\end{equation}
where $\mathrm{Proj}(\cdot)$ denotes the linear projection function and $\mathrm{TRN}_i(\cdot)$ denotes the $i$-th TRN module (composed of random clip sampling, feature concatenation and feature projection). For each temporal granularity, we align feature distributions across domains during temporal modeling of human cues. This is accomplished by a loss for the human encoder as follows:
\begin{equation}
\begin{aligned}
    \mathcal{L}^{\mathrm{hm}}
    & = \mathcal{L}^{ce}_{\mathrm{hm}} \left(\mathcal{H}\left(Z^{\mathrm{hm}}\right), y\right) \\
    & + \frac{\lambda^{\mathrm{hm}}}{M-1} \sum_{i=1}^{M-1} \mathcal{L}^{ce}_{\mathrm{trn}_i} \left(z_i^{\mathrm{hm}}, y_d\right) \\
    & + \frac{\lambda^{\mathrm{hm}}}{M-1} \sum_{i=1}^{M-1} \mathcal{L}^{ce}_{\mathrm{proj}} \left(\mathrm{Proj}(\hat{x}_i^{\mathrm{hm}}), y_d\right) ,
    \label{eq:hm-enc-loss}
\end{aligned}
\end{equation}
where $\lambda^{\mathrm{hm}}$ is a trade-off hyperparameter and $\mathcal{H}\left(\cdot\right)$ is the video-level feature aggregation function. $\mathcal{L}^{ce}_{*}(\cdot)$ indicates the cross-entropy loss with a classifier, where the subscript ``$_*$'' is used for distinguishing different classifiers\footnotemark[1].
\footnotetext[1]{``$\mathrm{hm}$'' indicates the action classifier for video-level human features, ``$\mathrm{proj}$'' indicates the domain discriminator for aligning the projected clip-level features, ``$\mathrm{trn}_i$'' indicates the domain discriminator for aligning the $i$-th order temporal dynamics.}
$y_d$ indicates the domain label, \ie, $y_d=0$ indicates the source domain and $y_d=1$ indicates the target domain. In Eq.~\eqref{eq:hm-enc-loss}, the first term is for source action classification\footnotemark[2], and the second and third terms are for domain alignment (following DANN~\cite{DBLP:journals/jmlr/GaninUAGLLML16}).
The above loss guides the human encoder to figure out the domain-invariant human cues.
\footnotetext[2]{The cross-entropy loss for action classification in Eq.~\eqref{eq:hm-enc-loss} is not applied on target videos due to the absence of target labels.}

\subsection{Action-Correlated Temporal Modeling for Contexts}
\label{subsec:ctx-enc}
Our proposed context encoder $\mathrm{CtxEnc}(\cdot)$ aims to model the temporal dynamics of contexts, as videos contain non-human context cues correlated to the performed actions (\eg, objects, scenes) apart from human cues.
Since there are abundant contexts in videos, a model could lose the focus on action-correlated contexts, due to the misguidance of some contexts irrelevant to the performing actions.
To solve this problem, we introduce context prototypes as extra tokens to guide context temporal modeling. We adopt a Transformer-like architecture with attention mechanisms, aiming at simultaneously action-correlated context mining and context temporal modeling.

Specifically, for the $i$-th clip, the set of clip-level non-human context features is denoted by $x_i^{\mathrm{ctx}} = \{x_{i,j} | m_{i,j}<\lambda_m, j=1,2,\dots,HW\}$. Given the context feature set $x_i^{\mathrm{ctx}}$, we conduct $K$-means clustering \cite{macqueen1967classification} over context features in the whole (source/target) domain and obtain $K$ context prototypes $C^{\mathrm{ctx}}=[c_1^{\mathrm{ctx}}, c_2^{\mathrm{ctx}}, ..., c_K^{\mathrm{ctx}}]^T\in\mathbb{R}^{K\times D}$ for each domain.
Before clustering, we train a source-only classifier for context features $x_i^{\mathrm{ctx}}$ and drop context features with high entropy (low confidence) since they are less correlated to actions.
As a result, these context prototypes summarize several types of action-correlated contexts in the domain, while some contexts irrelevant to actions are filtered out.
Considering both clip-level context features and context prototypes, the input of the context encoder consists of $Z_0^{\mathrm{ctx}}=[\hat{x}_1^{\mathrm{ctx}}, ..., \hat{x}_M^{\mathrm{ctx}}]\in\mathbb{R}^{M\times D_v}$ and $C_0^{\mathrm{ctx}}=[\hat{c}_1^{\mathrm{ctx}}, ..., \hat{c}_K^{\mathrm{ctx}}]^T\in\mathbb{R}^{K\times D_v}$, whose items are given as follows:
\begin{equation}
\begin{aligned}
    \hat{x}_i^{\mathrm{ctx}} = \mathrm{Proj}\left( \mathrm{AvgPool} \left(x_i^{\mathrm{ctx}}\right) \right) ,~~
    \hat{c}_i^{\mathrm{ctx}} = \mathrm{Proj}\left( c_i^{\mathrm{ctx}} \right) ,
    \label{eq:ctx-input}
\end{aligned}
\end{equation}
where $\mathrm{Proj}(\cdot)$ denotes a linear projection function\footnotemark[3].\footnotetext[3]{The functions $\mathrm{Proj}(\cdot)$ in Eq.~\eqref{eq:hm-trn} and Eq.~\eqref{eq:ctx-input} have different parameters. In Eq.~\eqref{eq:ctx-input}, we omit the position encoding to ease notations.}
The proposed context encoder consists of a stack of $L_e$ identical layers, and the context modeling of the $l$-th layer ($0< l\leq L_e$) is formulated as follows:
\begin{equation}
\begin{aligned}
    Z_l^{\mathrm{ctx}} & = \mathrm{FFN}\left( \mathrm{SelfAttn}\left( Z_{l-1}^{\mathrm{ctx}} \right) + \mathrm{CrossAttn}\left( Z_{l-1}^{\mathrm{ctx}}, C_{l-1}^{\mathrm{ctx}} \right) \right),
    \label{eq:ctx-enc}
\end{aligned}
\end{equation}
where the enhanced clip-level context feature $Z_{l}^{\mathrm{ctx}}$ denotes the output of the $l$-th layer (also the input of the $(l+1)$-th layer). Similarly, $C_{l-1}^{\mathrm{ctx}}$ denotes context prototype features enhanced by the $(l-1)$-th context encoder layer, which is obtained by the identical attention modules in Eq.~\eqref{eq:ctx-enc}. $\mathrm{FFN}(\cdot)$ denotes a position-wise feedforward module, $\mathrm{SelfAttn}(\cdot)$ denotes a self-attention module\footnotemark[4]\footnotetext[4]{The modules $\mathrm{SelfAttn}(\cdot)$ in Eq.~\eqref{eq:hm-input} and Eq.~\eqref{eq:ctx-enc} have different parameters.}, and $\mathrm{CrossAttn}(\cdot, \cdot)$ denotes a cross-attention module with the first term as the query.
Functionally speaking, the self-attention module serves for temporal modeling given video clips, while the cross-attention module serves for mining action-correlated contexts.
The cross-attention module forces the context encoder to attend on action-correlated context cues in each clip by aggregating information from similar context prototypes.
In this way, the context encoder would better exploit action-correlated contexts in further human-context interaction modeling. $Z^{\mathrm{ctx}}=Z_{L_e}^{\mathrm{ctx}}\in\mathbb{R}^{M\times D}$ is the output of the context encoder, \ie, the output of the last encoder layer.

To exploit domain-invariant context cues, we conduct feature alignment simultaneously with the above action-correlated context temporal modeling. Specifically, we conduct cross-domain alignment on the output features of each encoder layer, leading to the following loss:
\begin{equation}
\begin{aligned}
    \mathcal{L}^{\mathrm{ctx}} = \frac{\lambda^{\mathrm{ctx}}}{L_e} \sum_{l=1}^{L_e} \mathcal{L}^{ce}_{\mathrm{ctx}_l} \left( Z_l^{\mathrm{ctx}}, y_d\right),
    \label{eq:ctx-enc-loss}
\end{aligned}
\end{equation}
where $\lambda^{\mathrm{ctx}}$ is a trade-off hyperparameter and $\mathcal{L}^{ce}_{\mathrm{ctx}_l}$ denotes the cross-entropy loss for aligning output features of the $l$-th context encoder layer with a domain discriminator.

\subsection{Human-Context Interaction Modeling}
\label{subsec:hmctx-enc}
As discussed in Section~\ref{sec:introduction}, contexts alone could mislead the action recognition across domains, since an action could correlate with different types of contexts in different domains.
Recall that actions are performed by humans, thus only contexts that the humans interact with in videos should be considered, \eg, humans interact with bows when playing bow shooting.
Accordingly, we propose a human-context decoder $\mathrm{HCDec}(\cdot)$ to associate action-correlated contexts with humans, aiming at modeling domain-invariant human-context interaction for domain adaptive action recognition.

Our proposed human-context decoder is stacked on top of the human encoder and context encoder.
Since the outputs of the human encoder and context encoder are in different feature spaces, we adopt a Transformer-like architecture with cross-attention modules as core, which takes the human encoder output as the query and the context encoder output as the key/value.
The decoder consists of a stack of $L_d$ identical layers, and the $l$-th layer ($0< l\leq L_d$) is formulated as follows:
\begin{equation}
\begin{aligned}
    Z_l^{\mathrm{hc}} & = \mathrm{FFN}\left( \mathrm{CrossAttn}\left( Z_{l-1}^{\mathrm{hc}}, Z^{\mathrm{ctx}} \right) \right),
    \label{eq:hc-dec}
\end{aligned}
\end{equation}
where $Z_0^{\mathrm{hc}}=Z^{\mathrm{hm}}$ and $Z_{l}^{\mathrm{hc}}$ denotes the output of the $l$-th layer and also the input of the $(l+1)$-th layer.
$Z^{\mathrm{hc}}=Z_{L_d}^{\mathrm{hc}}$ is the output of the human-context decoder, \ie, the output of the last decoder layer. $\mathrm{FFN}(\cdot)$ denotes a position-wise feedforward module and $\mathrm{CrossAttn}(\cdot, \cdot)$ denotes a cross-attention module with the first term as query\footnotemark[5].
\footnotetext[5]{The modules $\mathrm{FFN}(\cdot)$ and $\mathrm{CrossAttn}(\cdot, \cdot)$ in Eq.~\eqref{eq:ctx-enc} and Eq.~\eqref{eq:hc-dec} have different parameters.}
In Eq.~\eqref{eq:hc-dec}, by leveraging the cross-attention modules, the human-context decoder models the interaction between human features and their related context features.
The human-context decoder embeds human information into interaction information by residual connections.
Then the decoder output $Z_l^{\mathrm{hc}}$ denotes human-centric features, which takes into account both human cues and human-context interaction cues.

For domain-invariant human-centric video feature learning, we take the human-centric features before aggregation for domain alignment at different temporal granularities. For each temporal granularity of each decoder layer, we align feature distributions across domains simultaneously with human-context interaction modeling, leading to the loss as follows:
\begin{equation}
\begin{aligned}
    \mathcal{L}^{\mathrm{hc}} = \frac{\lambda^{\mathrm{hc}}}{L_d(M-1)} \sum_{l=1}^{L_d} \sum_{i=1}^{M-1} \mathcal{L}^{ce}_{\mathrm{hc}_{l,i}} \left(z_{l,i}^{\mathrm{hc}}, y_d\right),
    \label{eq:hc-dec-loss}
\end{aligned}
\end{equation}
where $\lambda^{\mathrm{hc}}$ is a trade-off hyperparameter, $z_{l,i}^{\mathrm{hc}}$ is the output feature of the $l$-th decoder layer for the $(i+1)$-order temporal dynamics and $\mathcal{L}^{ce}_{\mathrm{hc}_{l,i}}(\cdot)$ indicates the cross-entropy losses applied with a domain discriminator for feature alignment.

\textbf{Overall Training and Inference.}
On top of the human-context decoder, we aggregate the human-centric features $Z^{\mathrm{hc}}$ (from all temporal granularities) as final video-level features. For guiding domain-invariant human-centric feature learning, the loss applied on the video-level features is given as follows:
\begin{equation}
\begin{aligned}
    \mathcal{L}^{\mathrm{video}}
    = \mathcal{L}^{ce}_{\mathrm{cls}} \left(\mathcal{H}(Z^{\mathrm{hc}}), y\right)
    + \lambda^{\mathcal{H}} \mathcal{L}^{ce}_{\mathcal{H}} \left(\mathcal{H}(Z^{\mathrm{hc}}), y_d\right),
    \label{eq:action-loss}
\end{aligned}
\end{equation}
where $\lambda^{\mathcal{H}}$ is a trade-off hyperparameter and $\mathcal{H}\left(\cdot\right)$ is the video-level feature aggregation function. $\mathcal{L}^{ce}_{\mathrm{cls}}(\cdot)$ and $\mathcal{L}^{ce}_{\mathcal{H}}(\cdot)$ indicate the cross-entropy losses applied with an action classifier and a domain discriminator, respectively. In Eq.~\eqref{eq:action-loss}, the first term is for source action classification\footnotemark[6] and the second term is for domain alignment following DANN~\cite{DBLP:journals/jmlr/GaninUAGLLML16}. During training, the overall loss of the proposed HCTransformer is given as follows:
\footnotetext[6]{The cross-entropy loss for action classification in Eq.~\eqref{eq:action-loss} is not applied on target videos due to the absence of target labels.}
\begin{equation}
\begin{aligned}
    \mathcal{L} = \mathcal{L}^{\mathrm{hm}} + \mathcal{L}^{\mathrm{ctx}} + \mathcal{L}^{\mathrm{hc}} + \mathcal{L}^{\mathrm{video}}.
    \label{eq:overall-objective}
\end{aligned}
\end{equation}
By leveraging the proposed decoupled architecture, we simultaneously conduct temporal modeling and cross-domain feature alignment for action recognition across domains.
During the inference phase, we use the classifier in $\mathcal{L}^{ce}_{\mathrm{cls}}(\cdot)$ on top of video-level feature $\mathcal{H}(Z^{\mathrm{hc}})$ for action recognition.

\section{Experiments}
\label{sec:experiment}

\subsection{Experimental Setups}
\subsubsection{Benchmarks}
We evaluate our model on three widely used benchmarks, namely UCF-HMDB \cite{DBLP:conf/iccv/ChenKAYCZ19}, Kinetics-NecDrone \cite{DBLP:conf/wacv/ChoiSCH20} and EPIC-Kitchens-UDA \cite{DBLP:conf/cvpr/MunroD20}.

\textbf{UCF-HMDB}
is the most widely used benchmark in domain adaptive action recognition, which covers 12 overlapped categories between UCF101 \cite{DBLP:journals/corr/abs-1212-0402} and HMDB51 \cite{DBLP:conf/iccv/KuehneJGPS11}. Each category in UCF-HMDB corresponds to one or more categories in the original UCF101 or HMDB51 dataset. Specifically, the videos in UCF101 are mostly captured from certain scenarios or similar environments, and the videos in HMDB51 are captured from unconstrained environments and different camera views. This benchmark includes two transfer tasks, \ie, UCF$\to$HMDB (``U$\to$H'') and HMDB$\to$UCF (``H$\to$U'').

\begin{table}[t]
\small
\caption{Comparison with the state-of-the-art methods on UCF-HMDB in terms of ACC (\%). For each method in the table, we list the adopted backbone and pre-training dataset and report the results on UCF$\to$HMDB (``U$\to$H'') and HMDB$\to$UCF (``H$\to$U'').}
\label{tab:ucf-hmdb}
\centering
\begin{tabular}{c | c | c | c | c }
\hline Method                                               & Backbone    & Pre-train     & U$\to$H       & H$\to$U \\
\hline
\hline Source-only                                          & R101        & ImgNet        & 71.7          & 73.9 \\
\hline DANN\cite{DBLP:journals/jmlr/GaninUAGLLML16}         & R101        & ImgNet        & 75.3          & 76.4 \\
\hline JAN\cite{DBLP:conf/icml/LongZ0J17}                   & R101        & ImgNet        & 74.7          & 79.4 \\
\hline AdaBN\cite{DBLP:journals/pr/LiWSHL18}                & R101        & ImgNet        & 72.2          & 77.4 \\
\hline MCD\cite{DBLP:conf/cvpr/SaitoWUH18}                  & R101        & ImgNet        & 73.9          & 79.3 \\
\hline TA3N\cite{DBLP:conf/iccv/ChenKAYCZ19}                & R101        & ImgNet        & 78.3          & 81.8 \\
\hline ABG\cite{DBLP:conf/mm/LuoHW0B20}                     & R101        & ImgNet        & 79.2          & 85.1 \\
\hline \textbf{Ours}                                        & R101        & ImgNet        & \textbf{80.8} & \textbf{87.4} \\
\hline
\hline Source-only                                          & BNIncep     & ImgNet        & 69.4          & 72.7 \\
\hline STCDA\cite{DBLP:conf/cvpr/SongZ0Y0HC21}              & BNIncep     & ImgNet        & 76.9          & 85.1 \\
\hline \textbf{Ours}                                        & BNIncep     & ImgNet        & \textbf{77.8} & \textbf{86.7} \\
\hline
\hline Source-only                                          & DeiT-B      & ImgNet        & 72.7          & 75.3 \\
\hline DANN\cite{DBLP:journals/jmlr/GaninUAGLLML16}         & DeiT-B      & ImgNet        & 77.4          & 78.6 \\
\hline TA3N\cite{DBLP:conf/iccv/ChenKAYCZ19}                & DeiT-B      & ImgNet        & 78.1          & 79.4 \\
\hline \textbf{Ours}                                        & DeiT-B      & ImgNet        & \textbf{81.9} & \textbf{82.8} \\
\hline
\hline Source-only                                          & Swin-B      & ImgNet        & 73.7          & 73.6 \\
\hline DANN\cite{DBLP:journals/jmlr/GaninUAGLLML16}         & Swin-B      & ImgNet        & 79.2          & 78.3 \\
\hline TA3N\cite{DBLP:conf/iccv/ChenKAYCZ19}                & Swin-B      & ImgNet        & 80.2          & 80.0 \\
\hline \textbf{Ours}                                        & Swin-B      & ImgNet        & \textbf{85.6} & \textbf{85.1} \\
\hline
\hline Source-only                                          & MViTv2-B    & ImgNet        & 73.0          & 76.4   \\
\hline DANN\cite{DBLP:journals/jmlr/GaninUAGLLML16}         & MViTv2-B    & ImgNet        & 76.0          & 78.6   \\
\hline TA3N\cite{DBLP:conf/iccv/ChenKAYCZ19}                & MViTv2-B    & ImgNet        & 76.9          & 78.9   \\
\hline \textbf{Ours}                                        & MViTv2-B    & ImgNet        & \textbf{81.1} & \textbf{82.8} \\
\hline
\hline Source-only                                          & I3D         & K400          & 80.3          & 88.8 \\
\hline TA3N\cite{DBLP:conf/iccv/ChenKAYCZ19}                & I3D         & K400          & 81.4          & 90.5 \\
\hline SAVA\cite{DBLP:conf/eccv/ChoiSSH20}                  & I3D         & K400          & 82.2          & 91.2 \\
\hline STCDA\cite{DBLP:conf/cvpr/SongZ0Y0HC21}              & I3D         & K400          & 81.9          & 91.9 \\
\hline ACAN\cite{xu2022aligning}                            & I3D         & K400          & 85.4          & 93.8 \\
\hline CoMix\cite{DBLP:conf/nips/SahooSPSD21}               & I3D         & K400          & 86.7          & 93.9 \\
\hline \textbf{Ours}                                        & I3D         & K400          & \textbf{93.6} & \textbf{95.6} \\
\hline
\hline Source-only                                          & 2S-I3D      & K400          & 82.8          & 90.7 \\
\hline MD-DMD\cite{DBLP:conf/mm/YinZCCJ22}                  & 2S-I3D      & K400          & 82.2          & 92.8 \\
\hline MM-SADA\cite{DBLP:conf/cvpr/MunroD20}                & 2S-I3D      & K400          & 84.2          & 91.1 \\
\hline STCDA\cite{DBLP:conf/cvpr/SongZ0Y0HC21}              & 2S-I3D      & K400          & 83.1          & 92.1 \\
\hline CMCF\cite{DBLP:conf/iccv/KimTZ0SSC21}                & 2S-I3D      & K400          & 84.7          & 92.8 \\
\hline CIA\cite{DBLP:conf/cvpr/YangHSS22}                   & 2S-I3D      & K400          & 89.7          & 93.2 \\
\hline \textbf{Ours}                                        & 2S-I3D      & K400          & \textbf{94.7} & \textbf{97.2} \\
\hline
\hline \blue{Source-only}                                   & \blue{MViTv1-B}    & \blue{K400}  & \blue{84.2}          & \blue{93.7}   \\
\hline \blue{DANN\cite{DBLP:journals/jmlr/GaninUAGLLML16}}  & \blue{MViTv1-B}    & \blue{K400}  & \blue{86.7}          & \blue{94.9}   \\
\hline \blue{TA3N\cite{DBLP:conf/iccv/ChenKAYCZ19}}         & \blue{MViTv1-B}    & \blue{K400}  & \blue{89.4}          & \blue{95.8}   \\
\hline \blue{\textbf{Ours}}                                 & \blue{MViTv1-B}    & \blue{K400}  & \blue{\textbf{94.2}} & \blue{\textbf{98.1}} \\
\hline
\hline \blue{Source-only}                                   & \blue{MViTv2-B}    & \blue{K400}  & \blue{86.1}          & \blue{94.2}   \\
\hline \blue{DANN\cite{DBLP:journals/jmlr/GaninUAGLLML16}}  & \blue{MViTv2-B}    & \blue{K400}  & \blue{88.9}          & \blue{95.4}   \\
\hline \blue{TA3N\cite{DBLP:conf/iccv/ChenKAYCZ19}}         & \blue{MViTv2-B}    & \blue{K400}  & \blue{90.3}          & \blue{96.0}   \\
\hline \blue{\textbf{Ours}}                                 & \blue{MViTv2-B}    & \blue{K400}  & \blue{\textbf{94.7}} & \blue{\textbf{98.8}} \\
\hline
\hline
\end{tabular}
\end{table}

\textbf{Kinetics-NecDrone}
is a challenging benchmark with a large domain gap between the source and target domains.
This benchmark consists of 7 categories, whose source videos are from Kinetics400 dataset \cite{DBLP:conf/cvpr/CarreiraZ17} and target videos are from NecDrone dataset \cite{DBLP:conf/wacv/ChoiSCH20,DBLP:conf/eccv/ChoiSSH20}. Videos from Kinetics400 are collected from YouTube, which are captured from various environments. Videos from NecDrone are captured by drones inside a school gym, therefore these videos in NecDrone have very similar backgrounds. The differences between Kinetics400 and NecDrone leads to a large gap across domains.

\textbf{EPIC-Kitchens-UDA}
is the most widely used large-scale benchmark in domain adaptive action recognition~\cite{DBLP:conf/cvpr/MunroD20}.
This benchmark uses a subset of EPIC-Kitchens-55~\cite{DBLP:conf/eccv/DamenDFFFKMMPPW18}, which is a large-scale dataset for egocentric action recognition. In EPIC-Kitchens-UDA, three largest kitchens, namely P08, P01 and P22, are selected to form three domains, denoted by D1, D2 and D3. In EPIC-Kitchens-UDA, 8 largest verb categories, which form 80\% of the training action segments for these domains, are selected for evaluation. Note that the actions are not human-centric but hand-centric in this benchmark. We adopt this benchmark in experiments for demonstrating the versatility of our proposed decoupled human-centric learning paradigm.

\subsubsection{Training and test protocols}
In our experiments, we strictly follow the standard experimental setups in previous works \cite{DBLP:conf/iccv/ChenKAYCZ19,DBLP:conf/mm/LuoHW0B20,DBLP:conf/wacv/ChoiSCH20,DBLP:conf/eccv/ChoiSSH20,DBLP:conf/aaai/PanCAN20} for fair comparison on all benchmarks. We evenly divide each video into $M$ segments for training and testing.
For UCF-HMDB, Kinetics-NecDrone and EPIC-Kitchens-UDA, we set the segment number as 5, 4, 5, respectively.
For a fair comparison, we use the save data augmentation strategy during training for each benchmark following previous works: we adopt temporal augmentation but no spatial augmentation on UCF-HMDB and Kinetics-NecDrone, and we adopt both temporal and spatial augmentations on EPIC-Kitchens-UDA.
We use accuracy (ACC) for performance evaluation.

\begin{table}[t]
\small
\caption{Comparison with the state-of-the-art methods on Kinetics-NecDrone in terms of ACC (\%). In the table, we use \dag~ to indicate a learnable backbone. In the ``Setting'' column, ``Unsupervised'' denotes the default unsupervised domain adaptation setting, and ``Semi-supervised'' denotes the semi-supervised domain adaptation setting with a small amount of target labels for training. }
\label{tab:kinetics-necdrone}
\centering
\begin{tabular}{c | c | c | c }
\hline Method                                               & Backbone        & Setting         & ACC (\%)       \\
\hline
\hline Source-only                                          & I3D             & Unsupervised    & 16.2          \\
\hline TA3N\cite{DBLP:conf/iccv/ChenKAYCZ19}                & I3D             & Unsupervised    & 28.1          \\
\hline \textbf{Ours}                                        & I3D             & Unsupervised    & \textbf{32.9}              \\
\hline
\hline Source-only                                          & I3D\dag         & Unsupervised    & 17.2          \\
\hline DANN\cite{DBLP:journals/jmlr/GaninUAGLLML16}         & I3D\dag         & Unsupervised    & 22.3          \\
\hline ADDA\cite{DBLP:conf/cvpr/TzengHSD17}                 & I3D\dag         & Unsupervised    & 23.7          \\
\hline TA3N\cite{DBLP:conf/iccv/ChenKAYCZ19}                & I3D\dag         & Unsupervised    & 32.5          \\
\hline SAVA\cite{DBLP:conf/eccv/ChoiSSH20}                  & I3D\dag         & Unsupervised    & 31.6          \\
\hline UDA\cite{DBLP:conf/wacv/ChoiSCH20}                   & I3D\dag         & Unsupervised    & 15.1          \\
\hline SSDA\cite{DBLP:conf/wacv/ChoiSCH20}                  & I3D\dag         & Semi-supervised & 32.0          \\
\hline CO2A\cite{DBLP:conf/wacv/CostaZROSM022}              & I3D\dag         & Unsupervised    & 33.2          \\
\hline \textbf{Ours}                                        & I3D\dag         & Unsupervised    & \textbf{38.6}              \\
\hline
\hline
\end{tabular}
\end{table}

\begin{figure*}[t]
\begin{center}
    \centering
    \includegraphics[width=1.0\linewidth]{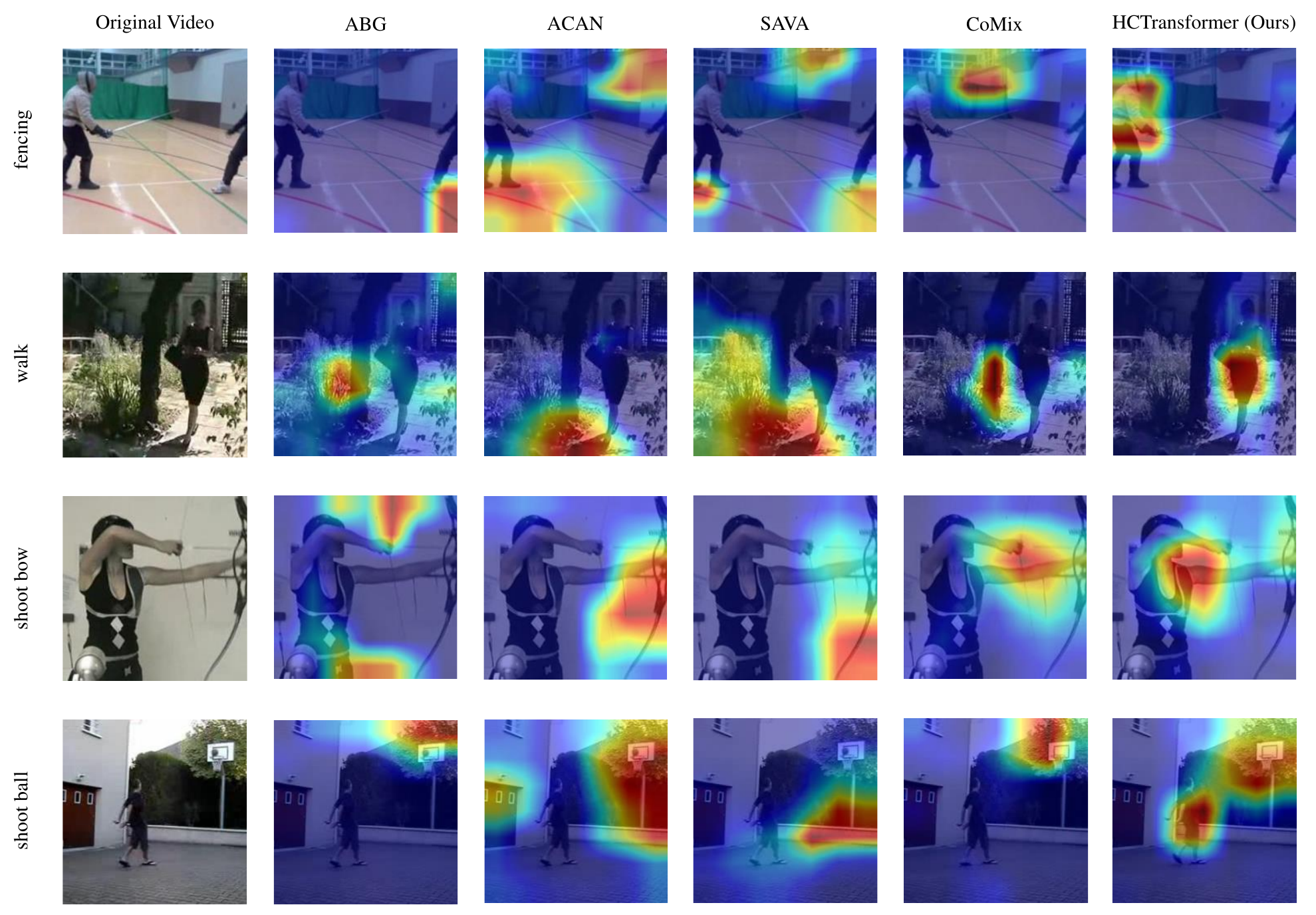}
\end{center}
\vskip -0.1in
\caption{
Grad-CAM visualization of more existing domain adaptive action recognition methods (in addition to TA3N~\cite{DBLP:conf/iccv/ChenKAYCZ19} as shown in Figure~\ref{fig:intro}), \ie, ABG~\cite{DBLP:conf/mm/LuoHW0B20}, ACAN~\cite{xu2022aligning}, SAVA~\cite{DBLP:conf/eccv/ChoiSSH20} and CoMix~\cite{DBLP:conf/nips/SahooSPSD21}.
For a clearer comparison, we also show the visualization results of our HCTransformer.
The results demonstrate that existing methods are prone to losing human cues in videos and our proposed HCTransformer focuses on human-centric action cues closely related to the performing actions.
Videos in the figure are from the target domain of UCF-HMDB. Best viewed in color.
}
\label{fig:more_baselines}
\end{figure*}

\subsubsection{Implementation details}
\label{subsubsec:implemental_details}
We mainly use ResNet101 (``R101'')~\cite{DBLP:conf/cvpr/HeZRS16}, BNInception (``BNIncep'') \cite{DBLP:conf/icml/IoffeS15} or I3D \cite{DBLP:conf/cvpr/CarreiraZ17} as backbone for clip feature extraction. The ResNet101 and BNInception backbones are pre-trained on ImageNet (``ImgNet'') \cite{DBLP:conf/cvpr/DengDSLL009}, and the I3D backbone is pre-trained on Kinetics-400 (``K400'') \cite{DBLP:conf/cvpr/CarreiraZ17}. The ResNet101 and BNInception backbones take a frame with $224\times224$ pixels as a clip for input, while the I3D backbone takes a clip of 16 frames with $224\times224$ pixels as input. We use average pooling for video-level feature aggregation on top of the human encoder and human-context decoder. If not specified, we fix backbone parameters and adopt $L_e=2$ context encoder layers and $L_d=2$ human-context decoder layers. We set the number of context prototypes $K=2N_{\mathrm{cls}}$. We set $D_v=512$, and $D=2048$ or $D=1024$ depending on the backbone. We omit the layer norm in the manuscript (including figures and texts), and we omit residuals of our Transformer architecture in formulas of Section \ref{sec:method}.
Following DANN~\cite{DBLP:journals/jmlr/GaninUAGLLML16}, all domain alignment losses follow an adversarial learning paradigm assisted by auxiliary domain discriminators and Gradient Reversal Layers~\cite{DBLP:journals/jmlr/GaninUAGLLML16}.

We use an off-the-self human detector and adopt a threshold $\lambda_m=0.5$ for partition pixels into human and non-human parts.
By default, we use a commonly used fully-convolutional network~\cite{DBLP:conf/cvpr/LongSD15} with a ResNet-50 backbone pretrained on COCO train2017~\cite{DBLP:conf/eccv/LinMBHPRDZ14} for detection.
For a video clip, if no humans are detected, we use the maxpooled human mask of the video, \ie, by applying maxpooling over $M$ human masks of the $M$ clips, which describes the range of human actions in the video for modeling temporal dynamics.
In domain adaptive action recognition benchmarks, humans always exist in action videos.
However, since the adopted human detector is not trained with human mask annotations on domain adaptive action recognition benchmarks, humans in videos could be missed by the detector.
Therefore, we use the average human mask over the training set, when no humans are detected in a whole video.
Since EPIC-Kitchens-UDA is hand-centric rather than human-centric, we use a hand detector to detect hands instead of a human detector, \ie, a DeepLabV3 model~\cite{DBLP:journals/corr/ChenPSA17} pretrained on EgoHands \cite{DBLP:conf/iccv/BambachLCY15} in the default case.

We randomly sample one clip from each segment for training and sample the center clip of each segment for test. For a fair comparison, we do not apply spatial augmentation on UCF-HMDB and Kinetics-NecDrone, while we apply random crops, scale jitters and random horizontal flips as TSN \cite{DBLP:conf/eccv/WangXW0LTG16} on EPIC-Kitchens-UDA. We sample 16 pairs of source and target videos for training, \ie, the batch size is 32. For training our decoupled architecture, we adopt a two-stage training strategy, \ie, first train the human encoder and then train the context encoder and human-context decoder with fixed human encoder.
In the first training stage, we adopt SGD optimizer with an initial learning rate of $0.01$ and a weight decay of 0.0005. In the second stage, we adopt Adam optimizer with an initial learning rate of $0.001$ and no weight decay. The learning rate decays after each iteration following DANN \cite{DBLP:journals/jmlr/GaninUAGLLML16}. As for hyperparameters, we set $\lambda^{\mathrm{hm}}=0.5$, $\lambda^{\mathrm{ctx}}=0.5$, $\lambda^{\mathrm{hc}}=0.25$ and $\lambda^{\mathcal{H}}=0.25$.

\begin{table*}[t]
\small
\caption{Comparison with the state-of-the-art methods on EPIC-Kitchens-UDA in terms of ACC (\%). In the table, all methods are trained and test based on only the RGB modality, except  MM-SADA\cite{DBLP:conf/cvpr/MunroD20} is trained with the guidance of Flow modality. }
\label{tab:epic-kitchens-uda}
\centering
\begin{tabular}{c | c | c  c  c  c  c  c | c}
\hline Method               & Backbone      & D3$\to$D1     & D2$\to$D1     & D1$\to$D3     & D2$\to$D3     & D1$\to$D2     & D3$\to$D2     & Average \\
\hline
\hline Source-only          & I3D           & 35.4          & 34.6          & 32.8          & 35.8          & 34.1          & 39.1          & 35.3    \\
\hline DANN\cite{DBLP:journals/jmlr/GaninUAGLLML16}
                            & I3D           & 38.3          & 38.8          & 37.7          & 42.1          & 36.6          & 41.9          & 39.2    \\
\hline ADDA\cite{DBLP:conf/cvpr/TzengHSD17}
                            & I3D           & 36.3          & 36.1          & 35.4          & 41.4          & 34.9          & 40.8          & 37.4    \\
\hline TA3N\cite{DBLP:conf/iccv/ChenKAYCZ19}
                            & I3D           & 40.9          & 39.9          & 34.2          & 44.2          & 37.4          & 42.8          & 39.9    \\
\hline CoMix\cite{DBLP:conf/nips/SahooSPSD21}
                            & I3D           & 38.6          & 42.3          & 42.9          & 49.2          & 40.9          & 45.2          & 43.2    \\
\hline MM-SADA\cite{DBLP:conf/cvpr/MunroD20}
                            & I3D           & 42.1          & 41.7          & 39.7          & 46.1          & 45.0          & 48.4          & 43.9    \\
\hline
\hline \textbf{Ours}
                            & I3D           & 43.2          & 46.2          & 45.0          & 46.4          & 43.3          & 47.6          & \textbf{45.3}    \\
\hline
\end{tabular}
\end{table*}

\begin{table}[t]
\small
\caption{
Comparison between our HCTransformer and state-of-the-art methods on UCF$\to$HMDB in terms of Human Ratio.
The Human Ratio
measures the average proportion of the overlapping regions between the thresholded Grad-CAM maps and ground-truth human bounding boxes.
A higher value of Human Ratio indicates a greater emphasis on human cues.
In this experiment, for all the methods, we set the thresholding coefficient as 0.5 for thresholding the Grad-CAM maps.
}
\label{tab:human-ratio}
\centering
\begin{tabular}{c | c }
\hline Method                                               & ~~Human Ratio (\%)~~ \\
\hline
\hline TA3N\cite{DBLP:conf/iccv/ChenKAYCZ19}                & 60.8  \\
\hline ABG\cite{DBLP:conf/mm/LuoHW0B20}                     & 61.9  \\
\hline SAVA\cite{DBLP:conf/eccv/ChoiSSH20}                  & 59.8  \\
\hline ACAN\cite{xu2022aligning}                            & 61.1  \\
\hline ~~~~CoMix\cite{DBLP:conf/nips/SahooSPSD21}~~~~         & 62.9  \\
\hline \textbf{Ours}                                        & \textbf{66.9}  \\
\hline
\end{tabular}
\end{table}

\subsection{Comparison with the State-of-the-arts}
We compare our proposed HCTransformer with two types of existing methods, namely traditional domain adaptation methods with TRN features (\eg, DANN \cite{DBLP:journals/jmlr/GaninUAGLLML16}, JAN \cite{DBLP:conf/icml/LongZ0J17}, etc) and video domain adaptation methods (\eg, TA3N \cite{DBLP:conf/iccv/ChenKAYCZ19}, SAVA \cite{DBLP:conf/eccv/ChoiSSH20}, etc). The experiment results are summarized in Table \ref{tab:ucf-hmdb}, \ref{tab:kinetics-necdrone} and \ref{tab:epic-kitchens-uda}.

As shown in these tables, due to the domain gaps, source-only models show limited action recognition abilities in the target domains (\eg, on Kinetics-NecDrone).
This is very different from the general action recognition tasks (\eg, Kinetics-400 classification),
where training and test videos follow the same distribution~\cite{DBLP:conf/cvpr/CarreiraZ17,DBLP:conf/cvpr/GuSRVPLVTRSSM18}.
These results highlight the importance of the domain adaptive action recognition task, where unlabeled videos from the target domain are accessible for training.
Also, these results demonstrate the importance of developing methods specifically tailored for this task.
Overall, our proposed HCTransformer is specifically designed for this task,
and it achieves the best performance on all these three benchmarks under fair settings.

On UCF-HMDB, when using I3D as the backbone, our model improves the performance by up to 6.9\% and 1.7\% over the state-of-the-arts in U$\to$H and H$\to$U, respectively. Specifically, our model based on I3D with single RGB modality input outperforms existing models based on two-stream I3D (``2S-I3D'') with both RGB and Flow modality as inputs. When using two-stream I3D\footnotemark[7], our model obtains further performance improvement by 1.1\% and 1.6\% over ours using I3D.
\footnotetext[7]{We adopt 2S-I3D as the backbone by fusing the score from RGB and Flow streams without cross-modality interaction during training.}
Also, we conduct experiments using advanced Vision Transformers as backbones, \ie, DeiT~\cite{DBLP:conf/icml/TouvronCDMSJ21}, Swin~\cite{DBLP:conf/iccv/LiuL00W0LG21} and MViTv2~\cite{DBLP:conf/cvpr/LiW0MXMF22} pretrained on ImageNet, \blue{as well as MViTv1~\cite{DBLP:conf/iccv/0001XMLYMF21} and MViTv2~\cite{DBLP:conf/cvpr/LiW0MXMF22} pretrained on Kinetics-400.}
In these cases, our superiority over the prevailing TA3N further verifies the effectiveness of our HCTransformer.

\begin{figure}[t]
\begin{center}
    \centering
    \includegraphics[width=0.9\linewidth]{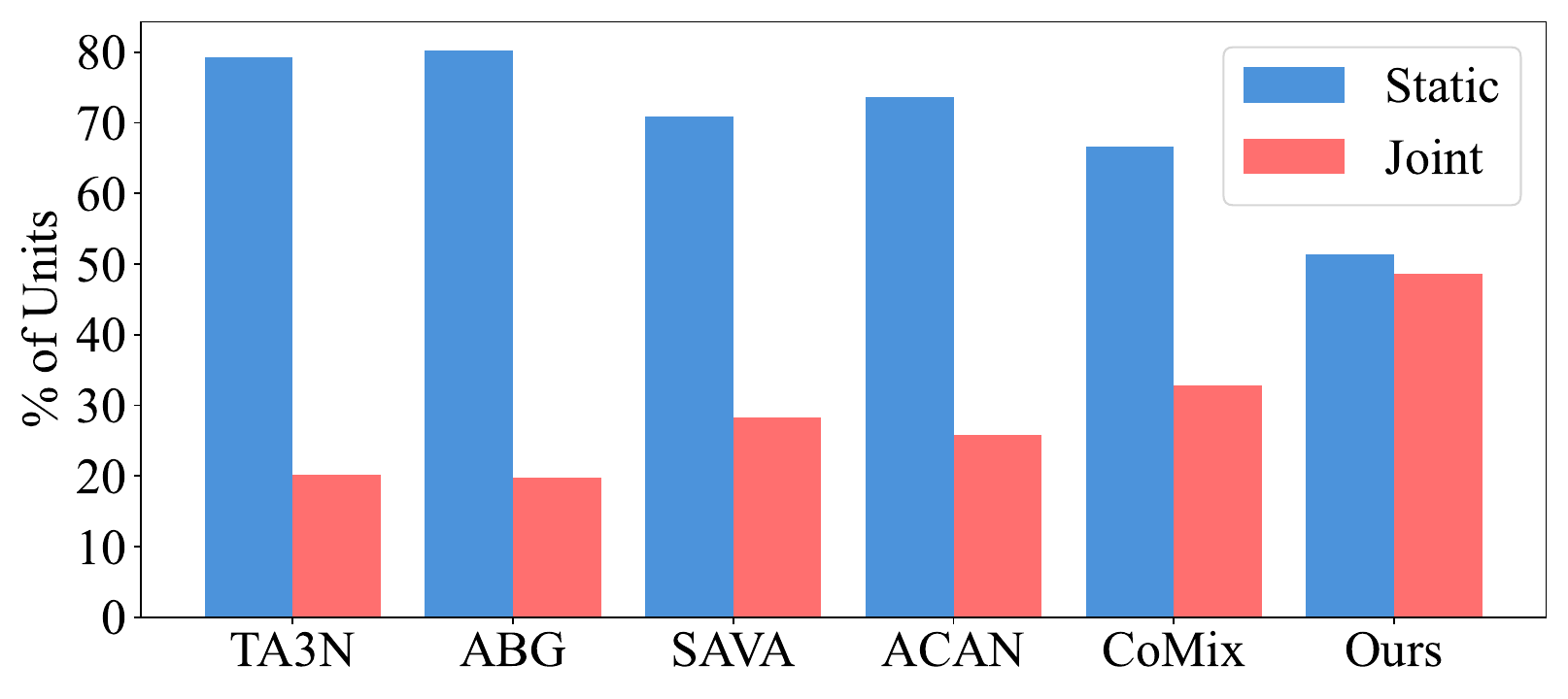}
\end{center}
\vskip -0.2in
\caption{
Quantitative analysis to the relative amount of static and dynamic information encoded by different domain adaptive action recognition models on UCF$\to$HMDB.
Following Kowal et al.~\cite{DBLP:conf/cvpr/KowalSIBWD22}, we use the unit-wise metric to quantify how many channels encode static/dynamic/joint information (``joint'' means both static and dynamic information are encoded).
For all the models, we use I3D as the backbone and use the features before classifier for analysis.
We empirically find that each domain adaptive action recognition model has two types of channels, namely static and joint ones.
Best viewed in color.
}
\label{fig:static_dynamic}
\end{figure}

\begin{figure*}[t]
\begin{center}
    \centering
    \includegraphics[width=0.9\linewidth]{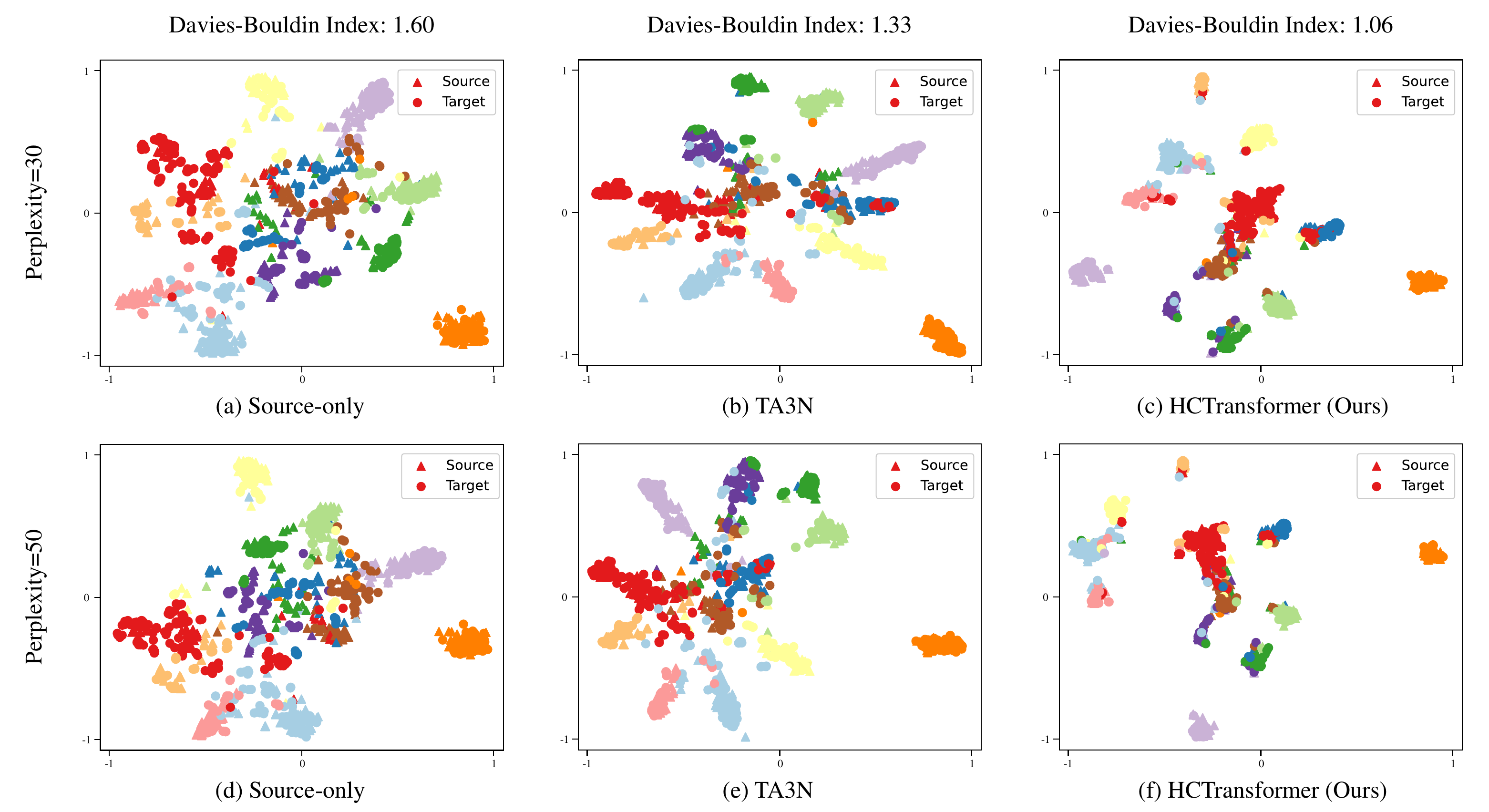}
\end{center}
\vskip -0.2in
\caption{
Feature distribution visualization by t-SNE~\cite{van2008visualizing} of the (a)/(d) source-only model, (b)/(e) TA3N and (c)/(f) our proposed HCTransformer on HMDB$\to$UCF.
In subfigures (a), (b) and (c), we use the default hyperparameter setting of scikit-learn with the perplexity as 30, and we set the perplexity as 50 in subfigures (d), (e) and (f).
For plots with the same perplexity, we keep all hyperparameters of t-SNE the same.
The triangle and circle markers denote the source and target video samples, and different colors denote different action classes.
In addition, we use a quantitative metric, namely the Davies-Bouldin Index~\cite{DBLP:journals/pami/DaviesB79}, to measure the separation between classes for each model in the original feature space.
A lower value of the Davies-Bouldin Index indicates better separation between classes.
Best viewed in color.
}
\label{fig:tsne}
\end{figure*}

On Kinetics-NecDrone, our HCTransformer with a fixed backbone achieves comparable performance with the state-of-the-art TA3N \cite{DBLP:conf/iccv/ChenKAYCZ19} and CO2A \cite{DBLP:conf/wacv/CostaZROSM022} based on learnable backbones.
When adopting a learnable backbone, our model obtains significant improvement over the state-of-the-arts, \ie, 5.4\% in terms of ACC over the state-of-the-art CO2A~\cite{DBLP:conf/wacv/CostaZROSM022}. In addition, our model trained in the unsupervised domain adaptation setting outperforms the SSDA~\cite{DBLP:conf/wacv/ChoiSCH20} trained in the semi-supervised domain adaptation setting (with extra 6\% target video labels), which demonstrates our superiority in model design.

Furthermore, our model achieves better performance compared with state-of-the-art methods on EPIC-Kitchens-UDA, which is hand-centric (rather than human-centric) consisting of egocentric action videos. Notably, without Flow modality guided and learnable I3D backbone, our HCTransformer can outperform MM-SADA~\cite{DBLP:conf/cvpr/MunroD20}.

The above experimental results demonstrate the superiority of our proposed HCTransformer over previous state-of-the-art methods, which attributes to the proposed decoupled human-centric learning paradigm.
In addition, the superior performance on EPIC-Kitchens-UDA further demonstrates the versatility of our proposed paradigm, \ie, our HCTransformer is effective for both human-centric and hand-centric domain adaptive action recognition.

\subsubsection{Comparison based on Grad-CAM}

In addition to the performance comparison with the state-of-the-art methods, we conduct a qualitative analysis by Grad-CAM visualization.
The results are given in Figure~\ref{fig:intro} and Figure~\ref{fig:more_baselines}.
As shown in the two figures, previous methods are prone to losing human cues in videos when recognizing actions across domains.
By contrast, our proposed HCTransformer focuses on human-centric action cues closely related to the performing actions in videos.

Moreover, we adopt a metric that measures the overlap between thresholded Grad-CAM maps and ground-truth human bounding boxes\footnotemark[8], which is termed the Human Ratio.
\footnotetext[8]{Following the pipeline of previous works~\cite{DBLP:conf/eccv/LinMBHPRDZ14,DBLP:conf/iccv/0001FWT21}, we annotate bounding boxes for the action performers in keyframes of validation videos from UCF$\to$HMDB.
These keyframes correspond to the center frames of video segments used during inference. }
The Human Ratio metric quantitatively measures how much the model depends on human cues in videos for recognizing action across domains.
A higher value of Human Ratio indicates a greater emphasis on human cues.
As shown in Table~\ref{tab:human-ratio}, our HCTransformer achieves a much higher value of Human Ratio compared with previous state-of-the-arts, which quantitatively demonstrates that our performance improvement is attributed to more concentration on humans when recognizing actions across domains.

\subsubsection{Analysis of static and dynamic information}

For a more comprehensive understanding of our HCTransformer, we conduct a quantitative analysis to static and dynamic information encoded by our model.
Fig.~\ref{fig:static_dynamic} presents the comparison with existing domain adaptive action recognition models in terms of the unit-wise metric proposed by Kowal et al.~\cite{DBLP:conf/cvpr/KowalSIBWD22}.
As shown in the figure, our HCTransformer involves much more joint channels compared with previous models, demonstrating that more dynamic information are encoded by our model.
This result indicates that our performance improvement is partially attributed to modeling more dynamic information of humans.

Inspired by the above analysis, we explore to extract more detailed dynamic information for further performance improvement.
To this end, we make a simple attempt, \ie, keeping the temporal dimension of the clip-level feature maps extracted by the I3D backbone.
Specifically, for the $i$-th clip, we first extract the basic clip-level feature map $\bar{x}_i\in\mathbb{R}^{2\times H\times W\times D}$ by the backbone, and then concatenate the features along the temporal dimension to produce the clip-level feature map $x_i\in\mathbb{R}^{H\times W\times 2D}$.
This clip-level feature map encodes more dynamic information, and our HCTransformer can conduct further human-centric feature learning based on this.
Our experiment results demonstrate that, with the I3D backbone, dropping the temporal pooling operation improves the performance by 0.8\% and 1.1\% on UCF$\to$HMDB and HMDB$\to$UCF, respectively.
The results indicate that exploring more detailed dynamic information may be a promising direction to improve domain adaptive action recognition in the future.

\subsubsection{Analysis to feature distribution}

\begin{figure*}[t]
\begin{center}
    \centering
    \includegraphics[width=0.95\linewidth]{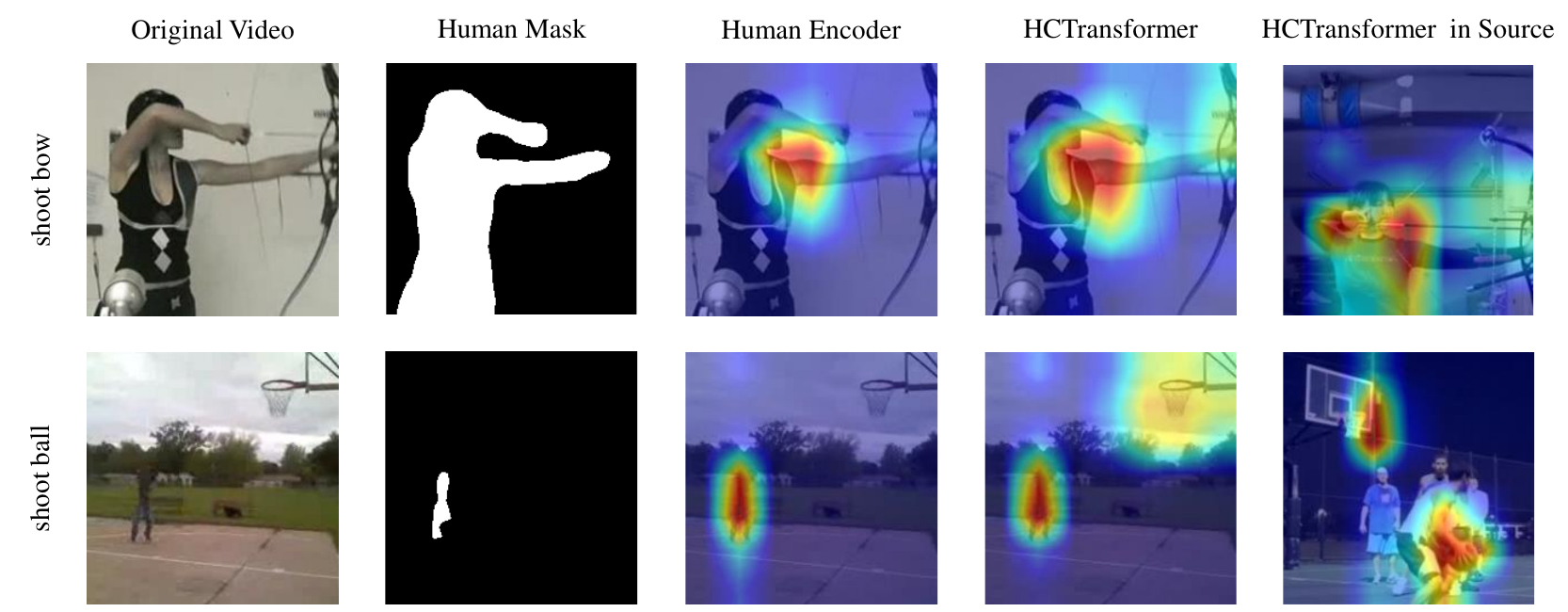}
\end{center}
\vskip -0.15in
\caption{Qualitative analysis of model components on UCF-HMDB. The two rows in the figure show visualization results of classes ``shoot bow'' and ``shoot ball'', respectively. In each row, we show the original target video, detected human mask, and Grad-CAM heatmaps of the proposed human encoder and full HCTransformer. In the last column of each row, we show Grad-CAM heatmaps of the full HCTransformer for a source video of the same class. As shown by the Grad-CAM heatmaps, our human encoder focuses on the human cues (\ie, body parts) in videos. By contrast, our HCTransformer focuses on both human cues and action-correlated context cues (\eg, bow, basket) in videos, which are domain-invariant across the source and target domains.
Best viewed in color.
}
\label{fig:gradcam}
\end{figure*}

\begin{table*}[t]
\small
\caption{Quantitative analysis of model components on UCF-HMDB, Kinetics-NecDrone and EPIC-Kitchens-UDA in terms of ACC (\%). In this table, ResNet101 is adopted as the backbone on UCF-HMDB, and I3D is adopted as the backbone on Kinetics-NecDrone and EPIC-Kitchens-UDA. ``HmEnc'' denotes the human encoder, ``CtxEnc'' denotes the context encoder, and ``HCDec'' denotes the human-context decoder.}
\label{tab:ablation}
\vskip -0.1in
\centering
\begin{tabular}{c | c  c | c | c}
\hline Method                   & ~~~U$\to$H~~       & ~~H$\to$U~~~       & ~Kinetics-NecDrone~   & ~EPIC-Kitchens-UDA~ \\
\hline
\hline Backbone+TRN             & ~77.2          & 80.6~          & 32.0          &42.0\\
\hline
       Backbone                 & ~71.1          & 75.1~          & 22.3          &39.2 \\
       Backbone+HmEnc           & ~76.7          & 83.2~          & 35.1          &43.5\\
       Backbone+CtxEnc          & ~72.8          & 70.9~          & 30.3          &39.2\\
       Backbone+HmEnc+CtxEnc    & ~77.2          & 83.5~          & 35.5          &43.8\\
       Full                     & ~80.8          & 87.4~          & 38.6          &45.3\\
\hline
       Full w/o Context Prototypes
                                & ~78.9          & 85.6~          & 37.3          &44.7\\
       ~~~~Full w/o Spatial Masking Module~~~~
                                & ~79.7          & 85.1~          & 35.4          &43.5\\
\hline
\end{tabular}
\end{table*}

In Figure~\ref{fig:tsne}, we compare our proposed HCTransformer with other methods by conducting t-SNE~\cite{van2008visualizing} visualization on HMDB$\to$UCF.
To conduct a more robust analysis to the feature distribution, we adopt two settings of hyperparamters for visualization.
First, we visualize the feature distribution of a source-only model to demonstrate the distribution shift between the source and target domains. As shown in Figure~\ref{fig:tsne} (a)/(d), the source and target distributions have many non-overlapping areas, \eg, points of different colors collide in central areas, and thus the classifier can hardly distinguish different classes well in the feature space. As shown in Figure~\ref{fig:tsne} (b)/(e), the prevailing TA3N~\cite{DBLP:conf/iccv/ChenKAYCZ19} mitigates the distribution shift, \eg, the \textcolor{rosy}{rosy}/\textcolor{skyblue}{sky blue} points of different shapes (domains) fall into a more compact cluster. However, there are still points of several (about six) colors crowded in central areas.
In contrast to TA3N, our proposed HCTransformer extracts features of less confusion between classes, \ie, points of different colors are better separated and are grouped into more compact clusters as shown in Figure~\ref{fig:tsne} (c)/(f).

Furthermore, we conduct a quantitative analysis to the feature distribution in the original feature space using the Davies-Bouldin Index~\cite{DBLP:journals/pami/DaviesB79}.
Specifically, the Davies-Bouldin Index measures a ratio between the intra-cluster distance and inter-cluster distance, and a lower value of Davies-Bouldin Index indicates better separation between different action classes.
As shown in Fig.~\ref{fig:tsne}, our proposed HCTransformer outperforms the prevailing TA3N in terms of Davies-Bouldin Index, which quantitatively verifies the effectiveness of our model.
Overall, both qualitative and quantitative analysis to the feature distribution demonstrate the effectiveness of our model.

\subsection{Analysis to Model Components}
In this part, we conduct extensive quantitative and qualitative experiments to analyze the effects of different model components.

\subsubsection{Quantitative ablation study}
First, we focus on the quantitative analysis, and results on UCF-HMDB, Kinetics-NecDrone and EPIC-Kitchens-UDA are summarized in Table \ref{tab:ablation}.
We apply an action classifier on top of backbone output features with DANN \cite{DBLP:journals/jmlr/GaninUAGLLML16} for feature alignment as our baseline (denoted by ``Backbone'').

\noindent \textbf{- Effect of human encoder.}
By stacking the human encoder (``HmEnc'') on top of the backbone for human-aware alignment, our model obtains significant improvement over the baseline, \eg, 5.6\% on U$\to$H and 8.1\% on H$\to$U.
Notably, our human encoder shows superiority over the vanilla Temporal Relation Network (TRN), \eg, on H$\to$U and Kinetics-NecDrone, which demonstrates the effectiveness of the proposed human-aware temporal modeling.

\noindent \textbf{- Effect of context encoder.}
By stacking our context encoder (``CtxEnc'') on top of the backbone, we find that introducing only the context encoder is inferior to introducing only the human encoder and may be even inferior to using the backbone alone in some cases (\eg, H$\to$U). This demonstrates that the human-centric action cues are more effective than non-human contexts for recognizing actions across domains. We also analyze the effect of context prototypes. According to ``Full'' vs. ``Full w/o Context Prototypes'', we find that the performance of HCTransformer drops when the context prototypes are removed, which demonstrates the effectiveness of the proposed action-correlated context modeling.

\noindent \textbf{- Effect of human-context decoder.}
By introducing both the context encoder (``CtxEnc'') and human-context decoder (``HCDec''), our proposed HCTransformer (``Full'') obtains substantial performance improvement over our human encoder (``Backbone+HmEnc''), \eg, 4.1\% on U$\to$H and 4.2\% on H$\to$U, which demonstrates the effectiveness of modeling human-context interaction. Also, we report the performance of combining the proposed human encoder and context encoder by late fusion~\cite{DBLP:conf/cvpr/KarpathyTSLSF14}, denoted by ``Backbone+HmEnc+CtxEnc''.
We find that using late fusion obtains very minor improvement compared with the human encoder alone (\eg, H$\to$U), which demonstrates the effectiveness of the proposed human-context interaction modeling by a query-based architecture.

\begin{figure}[t]
\begin{center}
    \centering
    \includegraphics[width=1\linewidth]{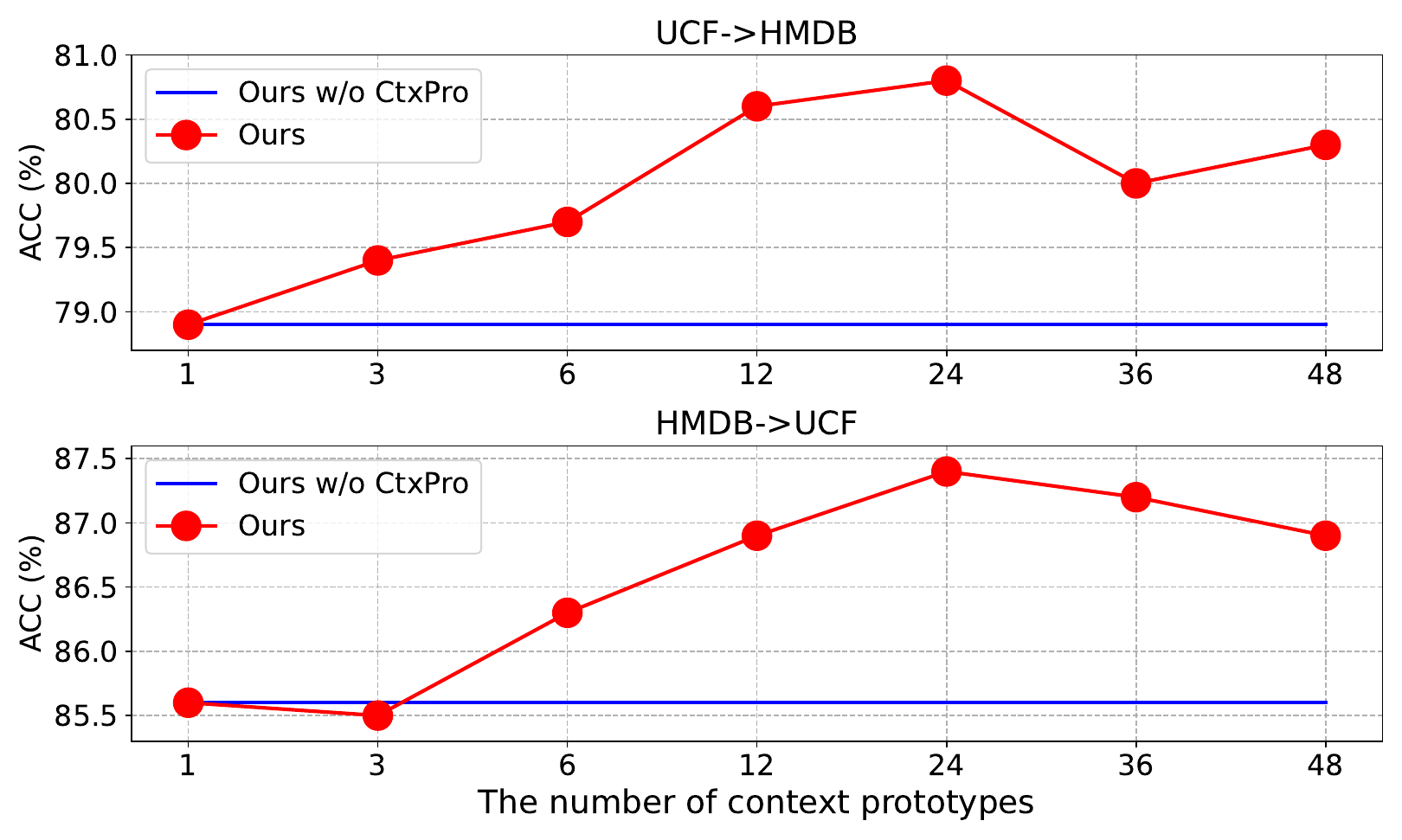}
\end{center}
\vskip -0.1in
\caption{Quantitative analysis on UCF-HMDB with different numbers of context prototypes. The ground-truth number of action classes in UCF-HMDB is 12.
The \textcolor{red}{red} and \textcolor{blue}{blue} lines denote our full HCTransformer (Ours) and our model without context prototypes in the human-context decoder (Ours w/o CtxPro), respectively.
Best viewed in color.}
\label{fig:cluster}
\end{figure}

\noindent \textbf{- Effect of spatial masking.}
Furthermore, we conduct experiments without the spatial masking module, where human masks are not introduced as prior knowledge for HCTransformer. For this variant, the clip-level feature maps are not divided into two parts, and each feature map is taken as inputs for both the human encoder and context encoder.
Although without the guidance of human masks, our proposed HCTransformer can still achieve state-of-the-art performance on the three benchmarks, which demonstrates the superiority of our architecture design for domain-invariant video feature learning.

Overall, the conclusions of quantitative ablation study on the three benchmarks are generally consistent, which demonstrates the effectiveness of our proposed model components. In addition, modeling human-context interaction beyond human cues is more effective on UCF-HMDB than that on Kinetics-NecDrone and EPIC-Kitchens-UDA, because UCF-HMDB involves more actions correlated with certain objects/scenes than the other two benchmarks.

\subsubsection{Qualitative ablation study}
We conduct a qualitative analysis on UCF-HMDB, and the results are shown in Figure~\ref{fig:gradcam}. As shown in the figure, our human encoder focuses on the body parts of action performers, according to the guidance of detected human masks.
Beyond the human encoder, our full model, namely HCTransformer, focuses on both action performers and action-correlated contexts in videos.
In addition, our proposed HCTrasformer focuses on regions of consistent semantics in videos across the source and target domains, which demonstrates the effectiveness of the proposed domain-invariant feature learning.
It should be noticed that our HCTransformer does not require perfect human masks.
For example, as shown in the second row of Figure~\ref{fig:gradcam}, the off-the-shelf human detector roughly extracts the outline of the human body with a loss of details, and the result shows that this human mask still effectively guides the proposed model for concentrating on human-centric cues.
In Figure~\ref{fig:intro}, we have shown our superiority over the prevailing TA3N~\cite{DBLP:conf/iccv/ChenKAYCZ19} by Grad-CAM examples, where our HCTransformer prefers to recognize determinant human-centric action cues and TA3N prefers to recognize vague contexts for domain adaptive action recognition.

\subsubsection{Analysis to context prototypes}

\begin{figure}[t]
\begin{center}
    \centering
    \includegraphics[width=1\linewidth]{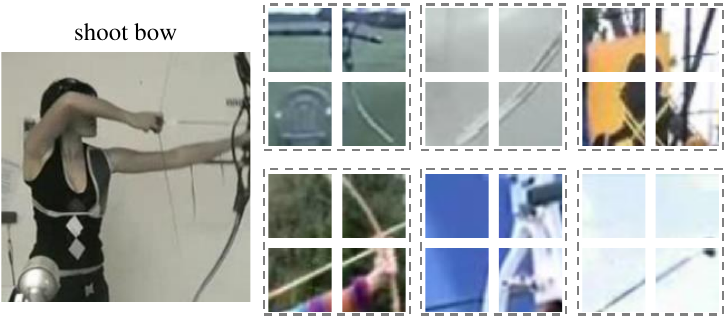}
\end{center}
\caption{Qualitative analysis of context prototypes. Left: A query video of the action ``shoot bow''.
Right: Patches from the most correlated cluster with the query ``shoot bow'' video.
We show six nearest patches to the queried prototype in the feature space.
For better visualization clarity, we demonstrate three extra adjacent patches for each patch (\ie, four adjacent patches as a group for demonstration).
As shown in the figure, the queried cluster mainly consists of bows and arrows that are closely correlated to the action ``shoot bow''.
In this experiment, we use a model trained on HMDB$\to$UCF for analysis, and the query video and patches in the figure are from HMDB.
Best viewed in color.
}
\label{fig:cluster_vis}
\end{figure}

First, we quantitatively analyze how the number of context prototypes affects the performance on UCF-HMDB, as shown in Figure~\ref{fig:cluster}.
In this experiment, we typically set the number of prototypes as multiples of the ground-truth number of action classes.
We find that our HCTransformer with a small number of context prototypes (\eg, $K=1$ or $K=3$) obtains comparable performance compared with our model without context prototypes. This is because that a small number of context prototypes (\eg, only one) cannot describe different categories of action-correlated contexts well in a domain.
When using a larger number of context prototypes (\eg, $K\geq 12$), our HCTransformer obtains substantial performance improvement over that without context prototypes.
In these cases, the context prototypes that summarize different categories of action-correlated contexts are introduced, and thus our model pays more attention on action-correlated contexts in context temporal modeling.
When using a very large number of context prototypes (\eg, $K=48$), HCTransformer does not obtain further performance improvement, and too many context prototypes introduce nonnegligible computation cost. Therefore, setting $K=2N_{cls}$ is an appropriate choice in practice according to this quantitative analysis.

\begin{figure}[t]
\begin{center}
    \centering
    \includegraphics[width=1\linewidth]{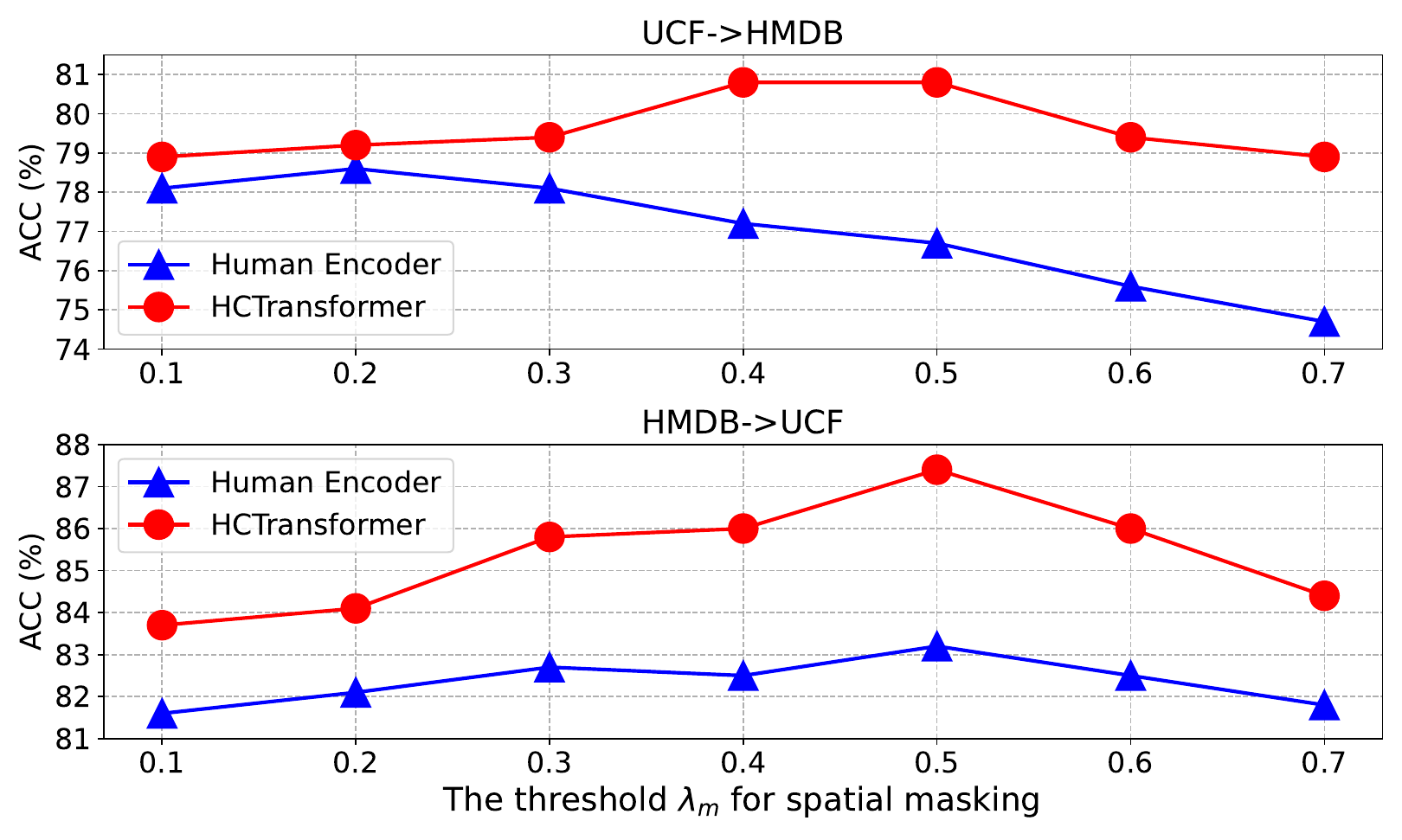}
\end{center}
\vskip -0.1in
\caption{Quantitative analysis on UCF-HMDB with different values of threshold $\lambda_m$ for spatial masking.
The \textcolor{red}{red} and \textcolor{blue}{blue} lines denote our full HCTransformer and our human encoder, respectively.
Best viewed in color.}
\label{fig:human_mask_threshold}
\end{figure}

Also, we show the patches in context clusters for qualitatively verifying the effectiveness of the proposed action-correlated context modeling. As shown in Figure~\ref{fig:cluster_vis}, we show six groups of nearest patches to the cluster prototype that is most correlated to the query ``shoot bow'' video. The results show that the queried cluster mainly consists of bows and arrows, which are closely correlated to the action ``shoot bow'' in the query video. Also, the cluster contains some patches from videos of other classes, \eg, the golf club at the bottom right. Although the cluster is confused with stick-like patterns of other classes, the overall semantics of the cluster would be disturbed very little, thus the classification performance is promised.

\subsubsection{Analysis to spatial masking}
On UCF-HMDB, we conduct a quantitative analysis of the threshold $\lambda_m$, which controls the partition of human and context feature sets in our proposed HCTransformer.
Generally speaking, a higher value of the threshold $\lambda_m$ indicates that fewer features are regarded as human cues by our model.
Notably, a very high threshold (\eg, $\lambda_m=0.7$) leads to no humans detected in many videos, and for these videos our model uses the average mask as illustrated in Section~\ref{subsubsec:implemental_details}. The quantitative results are summarized in Figure~\ref{fig:human_mask_threshold}, and we show the analysis below.

Firstly, the full HCTransformer always surpasses the human encoder when using different thresholds, which demonstrates the effectiveness of our proposed human-context interaction modeling beyond human-aware temporal modeling. In addition, the full HCTransformer obtains larger improvement over the human encoder when using a higher threshold $\lambda_m$. It is because a higher $\lambda_m$ leads to that more features are partitioned into the context set for human-context interaction modeling.

\begin{table}[t]
\small
\caption{
Quantitative analysis on UCF-HMDB and EPIC-Kitchens-UDA (EPIC) with different human/hand detectors in terms of ACC (\%).
Human detectors FCN~\cite{DBLP:conf/cvpr/LongSD15}, DeepLabV3~\cite{DBLP:journals/corr/ChenPSA17}, LR-ASPP~\cite{DBLP:conf/iccv/HowardPALSCWCTC19} and pose estimator Mask R-CNN~\cite{DBLP:conf/iccv/HeGDG17} are pretrained on COCO train2017~\cite{DBLP:conf/eccv/LinMBHPRDZ14},
and hand detectors DeepLabV3~\cite{DBLP:journals/corr/ChenPSA17}, HandBoxes~\cite{DBLP:conf/icb/ZhangZLSWL17}, and S3FD~\cite{DBLP:conf/iccv/ZhangZLSWL17} are pretrained on EgoHands~\cite{DBLP:conf/iccv/BambachLCY15}.
In each benchmark (subtable), we use the same threshold $\lambda_m$ for different detectors, and the first row shows the default detector.
}
\label{tab:humandet}
\vskip -0.0in
\begin{minipage}[t]{0.5\linewidth}
\hspace{0.05\linewidth}
\begin{tabular}{c | c | c}
\hline HumanDet                                             & U$\to$H   & H$\to$U \\
\hline
\hline FCN                                                  & 80.8      & 87.4  \\
\hline DeepLabV3                                            & 81.4      & 86.9  \\
\hline LR-ASPP                                              & 80.3      & 87.2  \\
\hline Mask R-CNN\footnotemark[9]                           & 80.3      & 86.5  \\
\hline
\end{tabular}
\end{minipage}
\hspace{0.1\linewidth}
\begin{minipage}[t]{0.45\linewidth}
\begin{tabular}{c | c }
\hline HandDet                                              & EPIC   \\
\hline
\hline DeepLabV3                                            & 45.3      \\
\hline HandBoxes                                            & 44.8      \\
\hline S3FD                                                 & 45.2      \\
\hline
\end{tabular}
\end{minipage}
\end{table}
\footnotetext[9]{When using Mask RCNN-based pose estimator for spatial masking, we first predict the keypoints and a
patch is considered as the human part if at least one keypoint is detected within it.}

Secondly, the performance of HCTransformer first increases and then decreases as the threshold $\lambda_m$ goes higher, and a medium threshold (\ie, $\lambda_m\in[0.3, 0.6]$) is generally effective on UCF-HMDB.
This result shows that it is important to make a good distinction between humans and contexts for the proposed decoupled human-centric learning paradigm.
When $\lambda_m$ is small (\eg, $\lambda_m=0.1$), many context cues are misclassified as humans.
When $\lambda_m$ is large (\eg, $\lambda_m=0.7$), many human cues are misclassified as contexts, and many videos would adopt the average mask that is mismatch to the ground-truth human masks in the videos.
When using a medium threshold, humans and contexts are well distinguished, and then the interaction between humans and contexts is well modeled by the adopted query-based architecture.

Thirdly, the performance of our human encoder shows different trends on the two transfer tasks, namely UCF$\to$HMDB and HMDB$\to$UCF. Specifically, the human encoder with a low threshold (\eg, $\lambda_m=0.2$) performs better than that with a relatively large threshold (\eg, $\lambda_m=0.5$) on UCF$\to$HMDB, which is inconsistent with the phenomenon on HMDB$\to$UCF. This attributes to the different properties of these two transfer tasks. Compared with HMDB$\to$UCF, recognizing contexts is more effective for domain adaptive action recognition on UCF$\to$HMDB.
Nevertheless, our full HCTransformer with a medium threshold significantly outperforms our human encoder with a low threshold on both transfer tasks, demonstrating the effectiveness of the proposed decoupled human-centric learning paradigm which models both human cues and human-context interaction cues.

To demonstrate the robustness of our proposed decoupled human-centric learning paradigm, we conduct experiments with different human/hand detectors for spatial masking on UCF-HMDB/EPIC-Kitchens-UDA.
As shown in Table~\ref{tab:humandet}, our HCTransformer obtains robust performance on both human-centric and hand-centric benchmarks,
\ie, the performance variation caused by using other human/hand detector is less than 0.9\% compared with the default detector.
It should be noted that, our HCTransformers with different human/hand detectors can outperform previous state-of-the-art methods, which demonstrates the effectiveness and robustness of our proposed decoupled human-centric learning paradigm.

\begin{figure}[t]
\begin{center}
    \centering
    \includegraphics[width=1\linewidth]{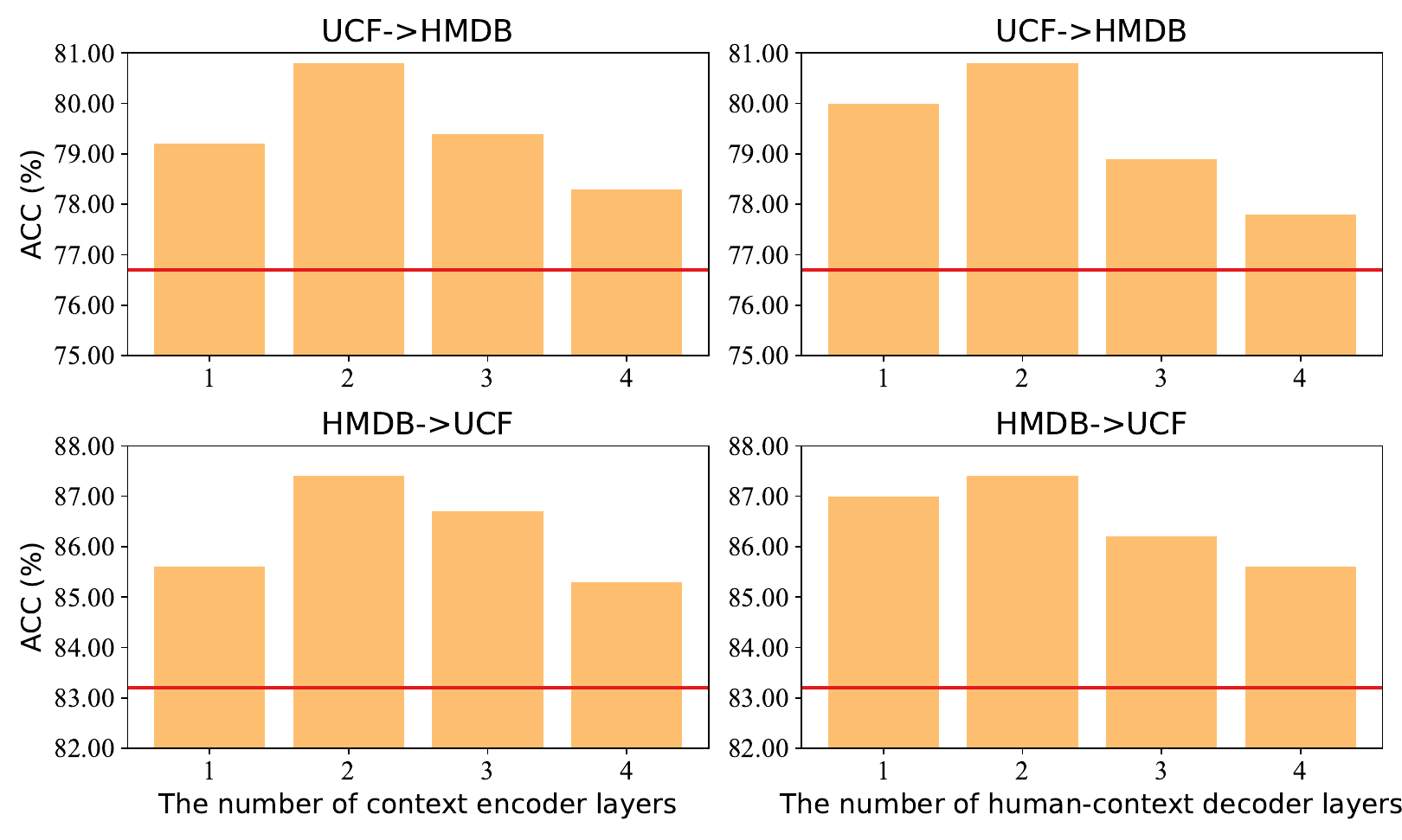}
\end{center}
\vskip -0.1in
\caption{Quantitative analysis on UCF-HMDB with different numbers of context encoder/human-context decoder layers. In each subfigure, the \textcolor{redline}{red} line represents the domain adaptive action recognition performance of our proposed human encoder. The number of context encoder/human-context decoder layers is fixed as two when varying the number of human-context decoder/context encoder layers.
Best viewed in color.}
\label{fig:number_layer}
\end{figure}

\subsubsection{Analysis to model depth}
On UCF-HMDB, we conduct a quantitative analysis of model depth, namely the number of layers of our proposed context encoder/human-context decoder. For simplicity, we term the context encoder as the encoder and the human-context decoder as the decoder in this quantitative experiment. Overall, our model obtains substantial performance improvement by stacking the context encoder and human-context decoder on top of the human encoder, as shown in Figure~\ref{fig:number_layer}.
From the figure, we find that adopting only one encoder/decoder layer obtains significant improvement over the baseline (our human encoder alone).
By increasing the number of encoder/decoder layers to two, we obtain better performance. Moreover, increasing the number of layers to three or four does not further improve the performance of domain adaptive action recognition, since too many parameters could easily make a model overfitted.

\section{Conclusion}
In this work, we reveal the importance of human-centric action cues for the domain adaptive action recognition task, which is intuitive but largely ignored or inadequately modeled by previous works.
This is because previous works mainly rely on the guidance of domain alignment losses, and our work is significantly different in imposing explicit constraints into the video feature extraction architecture for exploring the human-centric action cues.
We contribute a novel architecture named Human-Centric Transformer (HCTransformer), which explicitly decouples the video feature extraction process and mitigates the negative effects of action-agnostic contexts.
We conduct extensive experiments on three benchmarks, namely UCF-HMDB, Kinetics-NecDrone and EPIC-Kitchens-UDA, and our HCTransformer achieves state-of-the-art performance on all the three benchmarks.
\blue{
These experiment results verify that our HCTransformer can effectively improve the transferability of action recognition models by concentrating on human-centric cues through the proposed decoupled learning paradigm.
Moreover, our experiment results show that our HCTransformer is also effective for hand-centric actions though actions are usually human-centric.
These results demonstrate that our proposed HCTransformer is a versatile and promising solution for domain adaptive action recognition, as long as we learn the types of action subjects.
Overall, we believe that our work can provide insights for the field of human activity understanding, especially in the context of model robustness and trustworthiness.
}
In the future, we will explore more generalized and challenging transfer learning problems for action recognition, such as modality changes and action subject changes.


%

%

\ifCLASSOPTIONcompsoc
  \section*{Acknowledgments}
\else
  \section*{Acknowledgment}
\fi

The authors would like to thank Jia-Run Du, Yukun Qiu, Yi-Xing Peng, Haoxin Li, Ling-An Zeng, Jingke Meng, Zelin Chen and Zhilin Zhao for their valuable suggestions on model design, experiments or writing.
This work was supported partially by the National Key Research and Development Program of China (2023YFA1008503), NSFC (U21A20471), Guangdong NSF Project (No. 2023B1515040025, 2020B1515120085).

\ifCLASSOPTIONcaptionsoff
  \newpage
\fi



%

{
\bibliographystyle{IEEEtran}
\bibliography{vdaref}
}

\begin{IEEEbiography}[{\includegraphics[width=1in,height=1.25in,clip,keepaspectratio]{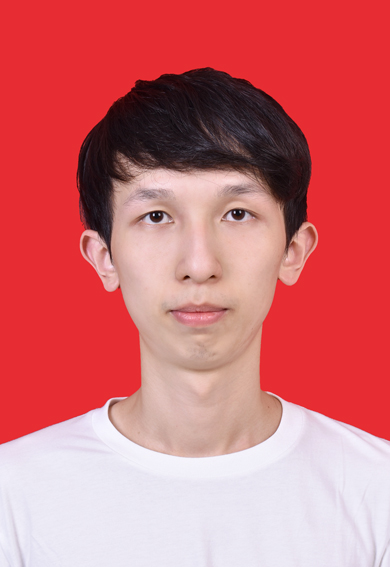}}]
{Kun-Yu Lin} received a B.S. and M.S. degree from the School of Data and Computer Science, Sun Yat-sen University, China. He is currently a PhD student in the School of Computer Science and Engineering, Sun Yat-sen University. His research interests include computer vision and machine learning.
\end{IEEEbiography}
\vskip -0.1in

\begin{IEEEbiography}[{\includegraphics[width=1in,height=1.25in,clip,keepaspectratio]{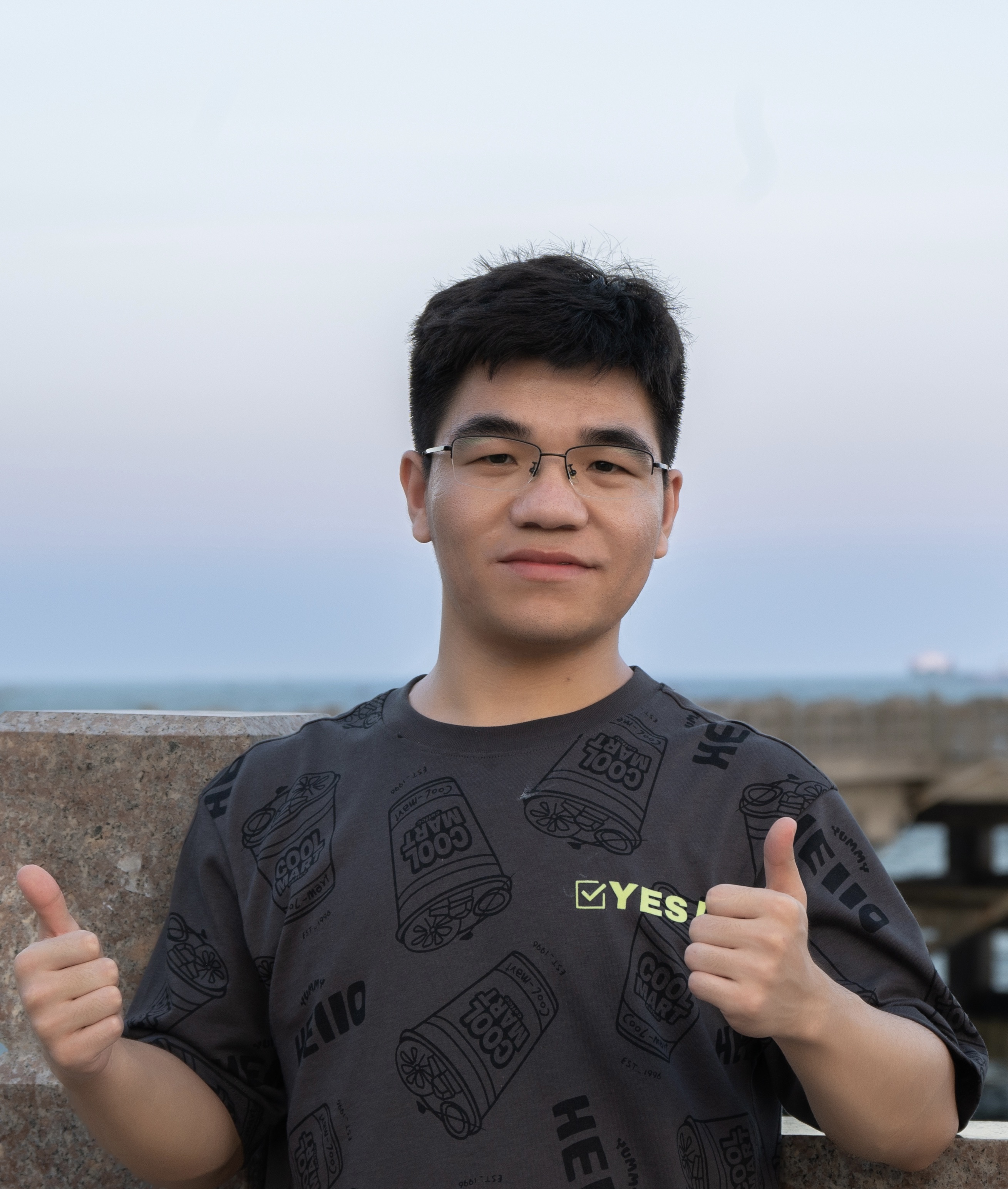}}]
{Jiaming Zhou} received a B.S. degree from School of Computer Science and Engineering, Sichuan University in 2020. He obtained a M.S. degree with the School of Computer Science and Engineering in Sun Yat-sen University. His research interests include computer vision and machine learning.
\end{IEEEbiography}
\vskip -0.1in

\begin{IEEEbiography}[{\includegraphics[width=1in,height=1.25in,clip,keepaspectratio]{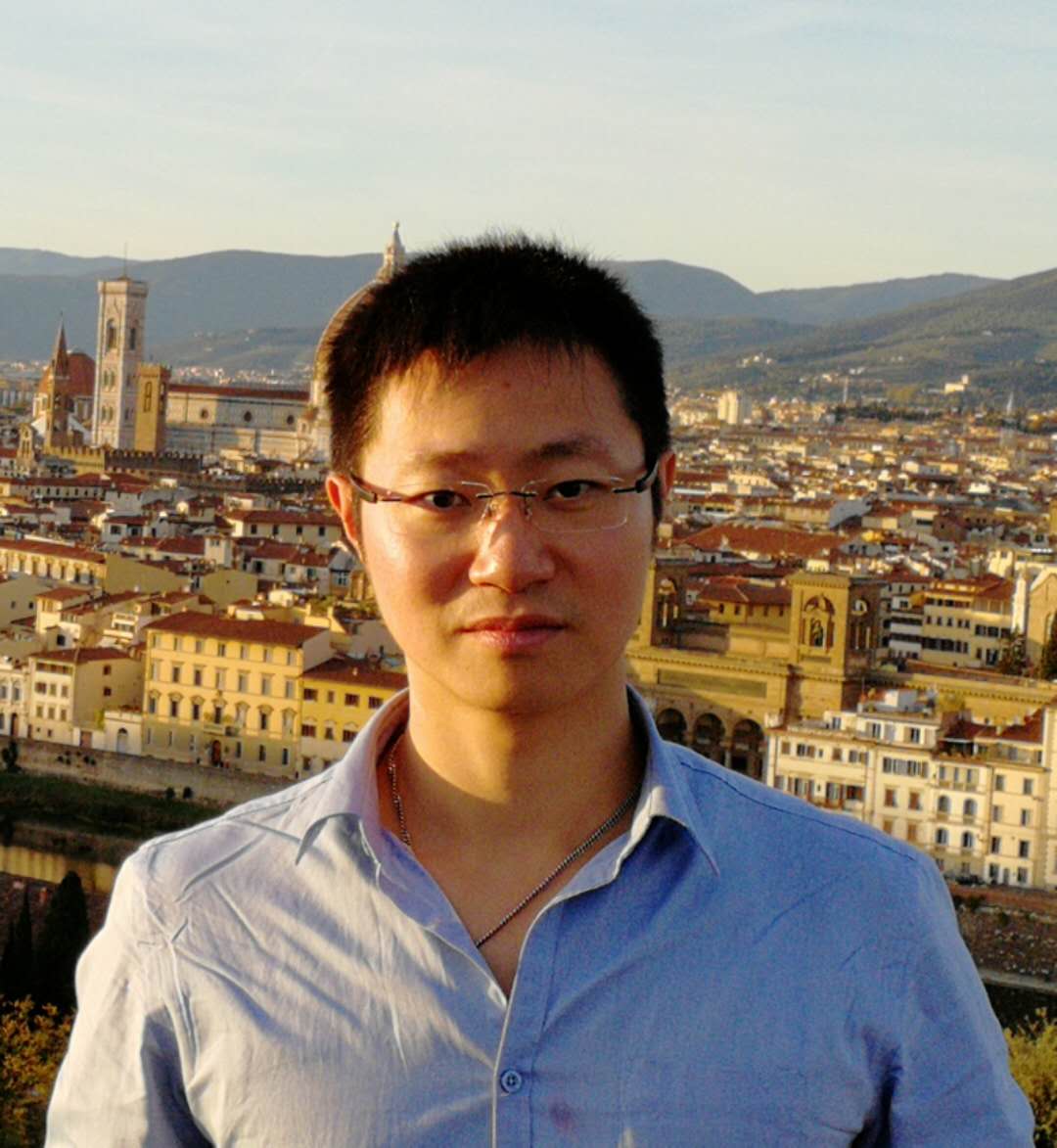}}]
{Dr. Wei-Shi Zheng} is now a Professor with Sun Yat-sen University. His research interests include person/object association and activity understanding in visual surveillance, and the related large-scale machine learning algorithm. He has ever served as area chairs of ICCV, CVPR, ECCV, NeurIPS, BMVC, IJCAI and AAAI. He is an associate editor of IEEE-TPAMI and Pattern Recognition. He has ever joined Microsoft Research Asia Young Faculty Visiting Programme. He is a Cheung Kong Scholar Distinguished Professor, a recipient of the Excellent Young Scientists Fund of the National Natural Science Foundation of China, and a recipient of the Royal Society-Newton Advanced Fellowship of the United Kingdom.
\end{IEEEbiography}


%








\end{document}